\documentclass{article}

\usepackage{arxiv}

\usepackage{microtype}
\usepackage{subfigure}
\usepackage{booktabs} 
\usepackage[export]{adjustbox}
\usepackage{algorithm}
\usepackage{mathtools}
\usepackage{amsthm}
\usepackage[numbers]{natbib}

\usepackage{amsmath,amsfonts,bm}

\def\eqref#1{(\ref{#1})}

\def\1{\bm{1}}

\DeclareMathAlphabet{\mathsfit}{\encodingdefault}{\sfdefault}{m}{sl}
\SetMathAlphabet{\mathsfit}{bold}{\encodingdefault}{\sfdefault}{bx}{n}

\newcommand{\norm}[1]{\left\lVert#1\right\rVert}

\newcommand{\proj}{\mathrm{proj}}

\usepackage{algorithm}
\usepackage{algorithmic}

\usepackage{hyperref}
\usepackage{url}

\usepackage[utf8]{inputenc}

\usepackage{microtype}
\usepackage{graphicx}
\usepackage{subfigure}
\usepackage{booktabs} 

\usepackage{hyperref}

\usepackage{caption}
\usepackage{cite}
\usepackage{amsmath,amssymb,amsfonts}
\usepackage{textcomp}
\usepackage{xcolor}

\theoremstyle{plain}
\newtheorem{theorem}{Theorem}[section]

\theoremstyle{definition}

\theoremstyle{remark}
\newtheorem{remark}[theorem]{Remark}

\title{Attacking c-MARL More Effectively: \\A Data Driven Approach}
\author{Nhan H. Pham$^{1}$, Lam M. Nguyen$^{1}$, Jie Chen$^{2}$, Hoang Thanh Lam$^{3}$ \\
\textbf{Subhro Das}$^{2}$, \textbf{Tsui-Wei Weng}$^{4}$\\
$^{1}$ IBM Research, Thomas J. Watson Research Center, Yorktown Heights, NY, USA\\
$^{2}$ MIT-IBM Watson AI Lab, IBM Research, Cambridge, MA, USA\\
$^{3}$ IBM Research, Dublin, Ireland\\
$^{4}$ Halıcıoğlu Data Science Institute, 
University of California San Diego, La Jolla, CA, USA. \\
\texttt{\{nhp,LamNguyen.MLTD\}@ibm.com},
\texttt{chenjie@us.ibm.com}, \texttt{t.l.hoang@ie.ibm.com},\\
\texttt{subhro.das@ibm.com}, \texttt{lweng@ucsd.edu}
}

\begin{document}

\maketitle

\begin{abstract}
In recent years, a proliferation of methods were developed for cooperative multi-agent reinforcement learning (c-MARL). However, the robustness of c-MARL agents against adversarial attacks has been rarely explored. In this paper, we propose to evaluate the robustness of c-MARL agents via a model-based approach, named \textbf{c-MBA}. Our proposed formulation can craft much stronger adversarial state perturbations of c-MARL agents to lower total team rewards than existing model-free approaches. In addition, we propose the first victim-agent selection strategy and the first data-driven approach to define targeted failure states where each of them allows us to develop even stronger adversarial attack without the expert knowledge to the underlying environment. Our numerical experiments on two representative MARL benchmarks illustrate the advantage of our approach over other baselines: our model-based attack consistently outperforms other baselines in all tested environments.
\end{abstract}

\section{Introduction}\label{sec_intro}
Deep neural networks are known to be vulnerable to adversarial examples, where a small and often imperceptible adversarial perturbation can easily fool the state-of-the-art deep neural network classifiers~\citep{szegedy2013intriguing,nguyen2015deep,goodfellow2014explaining, papernot2016limitations}. Since then, a wide variety of deep learning tasks have been shown to also be vulnerable to adversarial attacks, ranging from various computer vision tasks to natural language processing tasks \citep{jia2017adversarial,zhang2020adversarial,jin2020bert,alzantot2018generating}. 

Perhaps unsurprisingly, deep reinforcement learning (DRL) agents are also vulnerable to adversarial attacks, as first shown in \citep{huang2017adversarial} for atari games DRL agents. \citep{huang2017adversarial} study the effectiveness of adversarial examples on a policy network trained on Atari games under the situation where the attacker has access to the neural network of {the} victim policy. In \citep{lin2017tactics}, the authors further investigate a strategically-timing attack when attacking victim agents on Atari games at a subset of the time-steps. Meanwhile, \citep{kos2017delving} use the fast gradient sign method (FGSM) \citep{goodfellow2014explaining} to generate adversarial perturbation on the A3C agents \citep{mnih2016asynchronous} and explore training with random noise and FGSM perturbation to improve resilience against adversarial examples. While the above research endeavors focus on actions that take discrete values, another line of research {tackles} a more challenging problem on DRL with continuous action {spaces}~\citep{weng2019toward,gleave2019adversarial}. Specifically, \citep{weng2019toward} consider a two-step algorithm which determines adversarial perturbation to be closer to a targetted failure state using a learnt {dynamics model}, and \citep{gleave2019adversarial} propose a physically realistic threat model and demonstrate the existence of adversarial policies in zero-sum simulated robotics games. However, all the above works focused on the \textit{single} DRL setting.

\begin{figure*}[!ht]
    \begin{center}
    \includegraphics[width=0.95\linewidth,valign=t]{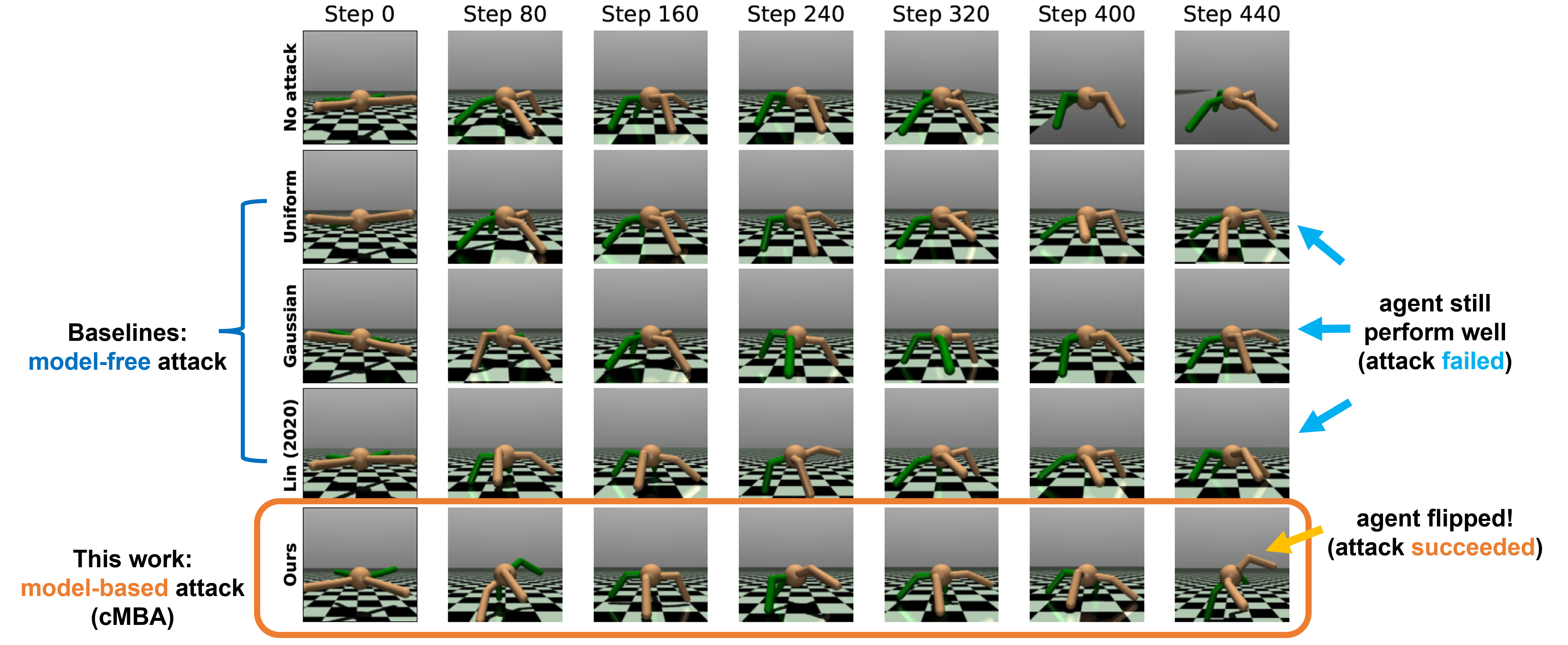}
    \caption{We illustrate the proposed model-based attack is powerful while other model-free attacks failed on attacking Agent $0$ in \textbf{Ant (4x2)} environment. The episode ends after 440 time steps under our \textbf{c-MBA} (the agent flips), demonstrating the effectiveness of our algorithm.}
    \label{fig:ant_linf_render}
    \end{center}
\end{figure*}

While most of the existing DRL attack algorithms focus on the \textit{single} DRL agent setting, in this work we propose to study the vulnerability of \textit{multi-agent} DRL, which has been widely applied in many safety-critical real-world applications including swarm robotics \citep{dudek1993taxonomy}, electricity distribution, and traffic control \citep{oroojlooyjadid2019review}. In particular, we focus on the collaborative multi-agent reinforcement learning (c-MARL) setting, where a group of agents is trained to generate joint actions to maximize the team reward. We note that c-MARL is a more challenging yet interesting setting than the \textit{single} DRL agent setting, as now one also needs to consider the interactions between agents, which makes the problem becomes more complicated.

Our contribution can be summarized as follows:  
\begin{itemize}\setlength\itemsep{-0.07ex}
    \item In this work, we propose the first \textit{model-based} adversarial attack framework on c-MARL, where we name it \textbf{c-MBA} (\textbf{M}odel-\textbf{B}ased \textbf{A}ttack on \textbf{c}-MARL). We formulate the attack into a two-step process and solve for adversarial state perturbation efficiently by existing proximal gradient methods. We show that our model-based attack is stronger and more effective than all of existing model-\textit{free} baselines. Besides, we propose a novel adaptive victim selection strategy and show that it could further increase the attack power of \textbf{c-MBA} by decreasing the team reward even more.
    \item To alleviate the dependence on the knowledge of the c-MARL environment, we also propose the first data-driven approach to define the targeted failure state based on the collected data for training the dynamics model. Our numerical experiments illustrate that \textbf{c-MBA} with the data-driven failure state is comparable and even outperforms \textbf{c-MBA} with the expert-defined failure state in many cases. Therefore, our data-driven approach is a good proxy to the optimal failure state when we have little or no knowledge about the state space of the c-MARL environments.
    
    \item We demonstrate on both multi-agent MuJoCo and multi-agent particle environments that our \textbf{c-MBA}
    consistently outperforms the SOTA baselines. We show that c-MBA can reduce the team reward up to $8-9\times$ when attacking the c-MARL agents. In addition, \textbf{c-MBA} with the proposed victim selection scheme matches or even outperforms other \textbf{c-MBA} variants in all environments with up to 80\% of improvement on reward reduction.
    
    
\end{itemize}

\textbf{Paper outline.} Section~\ref{sec:preliminary} discusses related works in adversarial attacks for DRL and present general background in c-MARL setting. We describe our proposed attack framework \textbf{c-MBA} in Section~\ref{subsec:cmba} for a fixed set of victim agents. In addition, we propose an alternative data-drive approach to determine the failure state for c-MBA in Section~\ref{subsec:data_drive_failure_state}. After that, we detail an adaptive strategy to design a stronger attack by selecting the most vulnerable victim agents in Section~\ref{sec:opt_adv_selection}. Section~\ref{sec:experiment} presents the evaluation of our approach on several standard c-MARL benchmarks. Finally, we summarize our results and future directions in Section~\ref{sec:conclusion}.

\section{Related work and background} \label{sec:preliminary}

\paragraph{Related work.}

Most of existing adversarial attacks on DRL agents are on \textit{single} agent \citep[for examples]{huang2017adversarial,lin2017tactics,kos2017delving,weng2019toward} while there are limited works that focus on the c-MARL setting. \citep{hu2022sparse} considers a different problem than ours where they want to find an optimally "sparse" attack by finding an attack with minimal attack steps. \citep{elhami2022adversarial} focuses on adversarial attacks against c-MARL system under \textit{time-delayed data transmission} setting which is not considered in this paper. \citep{guo2022towards} proposes a new robustness testing framework for c-MARL which considers state, action, and reward robustness. The design of state adversarial attack in \citep{guo2022towards} is based on gradient-based attack. \citep{lin2020robustness} proposes a two-step attack procedure to generate state perturbation for c-MARL setting which is the most relevant to our work. However, there are two major differences between our work and \citep{lin2020robustness}: (1) their attack is only evaluated under the StarCraft Multi-Agent Challenge (SMAC) environment \citep{samvelyan19smac} where the action spaces are discrete; (2) their approach is model-\textit{free} as they do not involve learning the dynamics of the environment and instead propose to train an adversarial policy for a fixed agent to minimize the the total team rewards. The requirement on training an adversarial policy is impractical and expensive compared to learning the dynamics model. To the best of our knowledge, there has not been any work considering adversarial attacks on the c-MARL setting using model-based approach on continuous action spaces. In this paper, we perform adversarial attacks on agents trained using MADDPG \citep{lowe2017multi} on two multi-agent benchmarks including multi-agent MuJoCo and multi-agent particle environments. Note that in the setting of adversarial attacks, once the agents are trained, policy parameters will be frozen and we do not require any retraining of the c-MARL agents during our attack.

\paragraph{Background in c-MARL.}
We consider multi-agent tasks with continuous action spaces modeled as a Decentralized Partially Observable Markov Decision Process (Dec-POMDP) \citep{oliehoek2016concise}. A Dec-POMDP has a finite set of agents $\mathcal{N}= \{1,\cdots,n\}$ associated with a set of states $\mathcal{S}$ describing global states, a set of continuous actions $\mathcal{A}_i$, and a set of individual state $\mathcal{S}_i$ for each agent $i \in \mathcal{N}$. Given the current state $s^i_t \in \mathcal{S}_i$, the action $a^i_t \in \mathcal{A}_i$ is selected by a parameterized policy $\pi^i: \mathcal{S}_i \to \mathcal{A}_i$. The next state for agent $i$ is determined by the state transition function $\mathcal{P}_i: \mathcal{S}\times \mathcal{A}_i \to \mathcal{S}$ and a new state $s_{t+1}^i \in \mathcal{S}_i$ is observed. Additionally, there is a reward function $\mathcal{R}: \mathcal{S}\times \mathcal{A} \to \mathbb{R}$ that generates the immediate reward the multi-agent system receives. In addition, Dec-POMDP is associated with an initial state distribution $\mathcal{P}_0$ and a discount factor $\gamma$. Training a c-MARL agent is to find a joint policy that maximize the total team rewards $\sum_{t} r_t$. Note that for the ease of exposition, we do not differentiate state and observation in this work and use them interchangeably throughout the paper.

\section{c-MBA: Model-based attack for c-MARL}\label{sec:method}



\subsection{Problem formulation and c-MBA attack}\label{subsec:cmba}
Our goal is to generate adversarial perturbations imposed to the victim agents' input (state) in order to deteriorate the total team reward. The added perturbations encourages the victim agents' state to be close to a desired failure state corresponding to low reward. To avoid sampling from the environment, we use a pre-trained model that learns the dynamics of the environment to predict the next state from the perturbed state and current action, then find the suitable noise that minimizes the distance between the predicted next state and a predefined target state. For now, we assume the target state is given and we show in Section~\ref{subsec:data_drive_failure_state} that this target state can actually be learned directly from the data. The overall attack can be formulated as an optimization problem as follows.

Formally, we consider a multi-agent setting with $|\mathcal{N}| = n$ agents, each agent $i \in \mathcal{N}$ receives state $s^i_t$ locally and takes action $a^i_t$ following the pre-trained c-MARL policy $\pi^{i}(s^i_t)$. Let $s_t = (s^1_t,\cdots,s^n_t) \in \mathcal{S}$ be the joint global state at time step $t$ which is concatenated from local states $s^i_t$ for each agent $i \in \mathcal{N}$. We also denote the joint action $a_t = (a^1_t,\cdots,a^n_t)$ concatenated from each agent's action $a^i_t$. Let $\mathcal{V}_t \subseteq \mathcal{N}$ be the set of victim agents at time step $t$, i.e. the set of agents that can be attacked. Let $f: \mathcal{S}\times \mathcal{A}\to \mathcal{S}$ be a parameterized function that approximates the dynamics of the environment, where $\mathcal{A}$ is the set of concatenated actions, one from each $\mathcal{A}_i$. Let $s_{fail}$ be the targeted failure state which corresponds to poor performance of the agent. We denote $\varepsilon$ as an upper bound on budget constraint w.r.t some $\ell_p$-norm $\norm{\cdot}_p$. The state perturbation $\Delta s = (\Delta s^1,\cdots,\Delta s^n)$ (we suppress the dependence on $t$ of $\Delta s$ to avoid overloading the notation) to $s_t$ is the solution to the following problem:
\begin{equation}\label{eq:sub_prob_1}
\begin{array}{rll}
\displaystyle\min_{\Delta s}~& d(\hat{s}_{t+1}, s_{fail})\\
\text{s.t.}~& \hat{s}_{t+1} = f(s_t, a_t)\\
& a^i_{t} = \pi^i(s^i_t + \Delta s^i),&\forall i \in \mathcal{N}\\
& \Delta s^i = \textbf{0},&\forall~i \notin \mathcal{V}_t\\
& \ell_{\mathcal{S}} \le s_t + \Delta s \le u_{\mathcal{S}}\\
& \norm{\Delta s^i}_p \le \varepsilon,&\forall i \in \mathcal{V}_t
\end{array}
\end{equation}
where $\textbf{0}$ is a zero vector, and the state vector follows a boxed constraint specified by $\ell_{\mathcal{S}}$ and $u_{\mathcal{S}}$.


Let us first provide some insights for the formulation \eqref{eq:sub_prob_1}. For each agent $i$, using the trained policy $\pi^i$, we can compute the corresponding action $a^i_t$ given its (possibly perturbed) local state $s^i_t$ or $s^i_t + \Delta s^i$. From the concatenated state-action pair $(s_t,a_t)$, we can predict the next state $\hat{s}_{t+1}$ via the learned dynamics model $f$. Then by minimizing the distance between $\hat s_{t+1}$ and the targeted failure state $s_{fail}$ subject to the budget constraint, we are forcing the victim agents to move closer to a damaging failure state in the next time step leading to low team reward.

Note that problem \eqref{eq:sub_prob_1} can be reformulated as a constrained nonconvex problem which can be efficiently solved by first-order method to obtain a stationary point. We defer the details to Sup. Doc.~\ref{app:solve_subproblem}. 
Finally, the full attack algorithm of c-MBA at timestep $t$ can be summarized in Alg.~\ref{alg:A1}.

\begin{algorithm}[hpt!]\caption{c-MBA algorithm at timestep $t$}\label{alg:A1}
\normalsize
\begin{algorithmic}[1]
   \STATE\label{step:i0}{\bfseries Initialization:} 
   \STATE\hspace{1.6ex} Given $s_t$, $s_{fail}$, $\pi$, $f$, $\mathcal{V}_t$; initialize $\Delta s = \varepsilon* sign(x)$ for $x\sim N(0,1)$, attack budget $\varepsilon$, $p$; choose learning rate $\eta > 0$
   \STATE\hspace{0ex}\label{step:o1}{\bfseries For $k = 0,\cdots, K-1$ do}
   \vspace{0.25ex}   
  \STATE\hspace{2ex}\label{step:o2} Compute $a_t = (a^1_t,\cdots,a^n_t)$ where $a^i_t = \pi^i(s_t^i + \Delta s_k^i)$ if $i \in \mathcal{V}_t$ and $a^i_t = \pi^i(s_t^i)$ otherwise.
  \STATE\hspace{2ex}\label{step:o3} Compute $\hat{s}_{t+1} = f(s_t,a_t)$.
  \STATE\hspace{2ex}\label{step:o4} Update $\Delta s_k$ using PGD.

   \STATE\hspace{0ex}{\bfseries End For}
\end{algorithmic}
\end{algorithm}

\textbf{Learning dynamics model.} One of the key enabler to solve \eqref{eq:sub_prob_1} is the availability of the learned dynamics model $f$. If the dynamics is known, we can solve \eqref{eq:sub_prob_1} easily with proximal gradient methods. However, in practice we often do not have the full knowledge of the environment and thus if we would like to solve  \eqref{eq:sub_prob_1}, we can learn the dynamics model via some function approximator such as neural networks. 


Learning the dynamics model can be formulated as solving the following optimization problem
\begin{equation}\label{eq:dynamic_model}
\min_{\phi}~\sum_{t \in \mathcal{D}} \norm{  f(s_t, a_t; \phi) - s_{t+1}}^2,
\end{equation}

where $\mathcal{D}$ is a collection of state-action transitions $\{(s_t, a_t,s_{t+1})\}_{t \in \mathcal{D}}$ and $s_{t+1}$ is the actual state that the environment transitions to after taking action $a_t$ determined by a given policy. In particular, we separately collect transitions using the pre-trained policy $\pi_{tr}$ and a random policy $\pi_{rd}$ to obtain $\mathcal{D}_{train}$ and $\mathcal{D}_{random}$. The motivation of using the random policy to sample is to avoid overfitting the dynamics model to the trained policy. Then the dataset $\mathcal{D}$ is built as $\mathcal{D}= \mathcal{D}_{train} \cup \mathcal{D}_{random}$. Since \eqref{eq:dynamic_model} is a standard supervised learning problem, the dynamics model $f$ can be solved by existing gradient-based methods. We describe the full process of training the dynamics model in Alg.~\ref{alg:A2} where the \texttt{GradientBasedUpdate} step could be any gradient-descent-type update. In our experiments, we notice that the quality of the dynamics model does not significantly affect the result as seen in Figure~\ref{fig:ant_model_acc}. We find that even when we use a less accurate dynamics model, our proposed c-MBA attack is still effective. Our attack using the dynamics model trained with only $0.2$ M samples for only 1 epoch is comparable with ones using more accurate dynamics model (1M for 1 and 100 epochs) in the \textbf{Ant (4x2)} environment under both $\ell_\infty$ and $\ell_1$ budget constraint. We describe the whole process of training the dynamics model in Algorithm~\ref{alg:A2}.

\begin{algorithm}[hpt!]\caption{Training dynamics model}\label{alg:A2}
\normalsize
\begin{algorithmic}[1]
  \STATE\label{step:i0}{\bfseries Initialization:} Given pre-trained policy $\pi_{tr}$ and a random policy $\pi_{rd}$; initialize dynamics model parameter $\phi_0$.
  \STATE\hspace{0ex}\label{step:o1}Form $\mathcal{D}= \mathcal{D}_{train} \cup \mathcal{D}_{random}$ by collecting a set of transitions $\mathcal{D}_{train}$ and $\mathcal{D}_{random}$ using policy $\pi_{tr}$ and $\pi_{rd}$, respectively.
  \STATE\hspace{0ex}\label{step:o2}{\bfseries For $k = 0,1,\cdots $ do}
  \begin{equation*}
      \phi_{k+1} = \texttt{GradientBasedUpdate}(\mathcal{D},\phi_k)
  \end{equation*}
  \STATE\hspace{0ex}{\bfseries End For}
\end{algorithmic}
\end{algorithm}

\paragraph{Discussion: difference between the baseline \citep{lin2020robustness}.}
We note that the most closely related to our work is \citep{lin2020robustness}, where they also propose an attack algorithm to destroy c-MARL. However, there are two major differences between their approach and ours: (1) Their method is a \textit{model-free} approach based on training extra adversarial policies, which could be impractical as it requires a lot of samples and it may be difficult to collect the required "bad" trajectories to minimize the team reward (this requires the full control of all the agents in the c-MARL setting which may not be available in practice). On the other hand, our c-MBA is a \textit{model-based} approach, where we only need to have a rough surrogate of the environment. This is an more practical scenario, and even very crude dynamics model could make c-MBA effective (see \textbf{Experiment (III)} in Section~\ref{sec:experiment}). (2) They did not leverage the unique setting in MARL to select most vulnerable agent to attack, while in the next section, we show that with the victim agent selection strategy, we could make c-MBA an even stronger attack algorithm (also see \textbf{Experiment (IV)} in Section~\ref{sec:experiment} for more details).

\subsection{Learning failure states from data}\label{subsec:data_drive_failure_state}

In order to solve \eqref{eq:sub_prob_1}, we need to specify the failure state $s_{fail}$ which often requires prior knowledge of the state definition of the c-MARL environment. To make our method more flexible, we propose the first data-driven approach to learn the failure state. As our c-MBA attack involves training a dynamics model of the environment by collecting the transition data, we can learn the failure state from directly the pre-collected dataset without extra overhead. Based on the observation that the a failure state should be a resulting state where the reward corresponding to that transition is low, we sort the collected transition $(s_t,a_t,\hat{s}_{t},r_t)$ by ascending order of the reward $r_t$ and choose the failure state to be $\hat{s}^i_{s}$ that corresponds to the lowest $r_t$ in the dataset. This process is described in Algorithm~\ref{alg:A1_2}.

\begin{algorithm}[hpt!]\caption{Learning failure state from collected data}\label{alg:A1_2}
\normalsize
\begin{algorithmic}[1]
   \STATE\label{step:i0}{\bfseries Input:}: a dataset $\mathcal{D} = \{(s_t,a_t,s_{t+1})\}_{t\in\mathcal{D}}$ as a collection of transitions.
   \STATE\hspace{0ex} Sort the transition by ascending order of reward $r_t$
   \vspace{0.25ex}   
  \STATE\hspace{0ex} Determine $(s_{(1)},a_{(1)},\hat{s}_{(1)},r_{(1)})$ as the transition corresponding to the minimum reward.
  \STATE\hspace{0ex} Set $s_{fail} = \hat{s}_{(1)}$.
\end{algorithmic}
\end{algorithm}

Using this data-driven strategy, we show that performing c-MBA attack with the learned failure state either matches or performs better than the expert-defined failure state for multi-agent MuJoCo environments. In addition, the data-driven approach demonstrates its advantage in the multi-agent particle environment where we do not have expert knowledge of the state space. Overall, c-MBA using the learned failure state shows its superior performance over other model-free baselines in most cases in the experiments.

\subsection{Crafting stronger attack with c-MBA -- victim agent selection strategy}\label{sec:opt_adv_selection}

In this subsection, we propose a new strategy to select most vulnerable victim agents with the goal to further increase the power of our cMBA. We note that this scenario is unique in the setting of \textit{multi-agent} DRL setting, as in the \textit{single} DRL agent setting we can only attack the same agent all the time. To the best of our knowledge, our work is the first to consider the victim selection strategy as \citep{lin2020robustness} only use one fixed agent to perform the attack. As a result, we can develop a stronger attack by selecting appropriate set of "vulnerable" agents. This strategy can be effective in the "sparse attack" setting when only a few agents in the team are attacked \citep{hu2022sparse}. To start with, we first formulate a mixed-integer program to perform the attack on a set of victim agents as below:
\begin{equation}\label{eq:opt_adv_prob_1}
\begin{array}{lll}
\min_{\Delta s, w}~& d(\hat{s}_{t+1}, s_{fail})\\
\text{s.t.}~& \hat{s}_{t+1} = f(s_t, a_t)\\
& a_{t}^i = \pi_i(s_t^i + w_i\boldsymbol{\cdot}\Delta s^i),&\forall i \in \mathcal{N}\\
& \ell_{\mathcal{S}} \le s_t^i + \Delta s^i \le u_{\mathcal{S}},&\forall i \in \mathcal{N}\\
& \norm{\Delta s^i}_p \le \varepsilon,&\forall i \in \mathcal{N}\\
& w_i \in \{0,1\},&\forall i \in \mathcal{N}\\
& \sum_{i} w_i = n_v
\end{array}
\end{equation}
where we introduce a new set of binary variables $\{w_i\}$ to select the suitable agent to attack. 

Due to the existence of the new binary variables, problem \eqref{eq:opt_adv_prob_1} is much harder to solve. Therefore, we instead propose to solve a proxy of \eqref{eq:opt_adv_prob_1} as follows
\begin{equation}\label{eq:opt_adv_prob_2}
\begin{array}{lll}
\min_{\Delta s, \theta}~& d(\hat{s}_{t+1}, s_{fail})\\
\text{s.t.}~& \hat{s}_{t+1} = f(s_t, a_t)\\
& a_{t}^i = \pi^i(s_t^i + W_i(s_t;\theta)\boldsymbol{\cdot}\Delta s^i),&\forall i \in \mathcal{N}\\
& \ell_{\mathcal{S}} \le s_t^i + \Delta s^i \le u_{\mathcal{S}},&\forall i \in \mathcal{N}\\
& \norm{\Delta s^i}_p \le \varepsilon,&\forall i \in \mathcal{N}\\
& 0 \le W_i(s_t;{\theta}) \le 1,&\forall i \in \mathcal{N}
\end{array}
\end{equation}
where $W(s;\theta): \mathcal{S} \rightarrow \mathbb{R}^n$ is a function parametrized by $\theta$ that takes current state $s$ as input and returns the weight to distribute the noise to each agent. Suppose we represent $W(s;\theta)$ by a neural network, we can rewrite the formulation \eqref{eq:opt_adv_prob_2} as \eqref{eq:opt_adv_prob_3} because the last constraint in \eqref{eq:opt_adv_prob_2} can be enforced by using a softmax activation in the neural network $W(s;\theta)$ 
\begin{equation}\label{eq:opt_adv_prob_3}
\begin{array}{ll}
\min_{\Delta s, \theta}~& d(f(s_t, \pi(\{s_t^i + W_i(s_t;\theta)\cdot\Delta s^i\}_i)), s_{fail})\\
\text{s.t.}~& \Delta s \in \mathcal{C}_{p,\varepsilon,t} 
\end{array}
\end{equation}
where the notation $\pi(\{s_t^i + W_i(s_t;\theta)\cdot\Delta s^i\}_i)$ denotes the concatenation of action vectors from each agent to form a global action vector $[\pi^1(s_t^1 + W_1(s_t; \theta)\cdot\Delta s^1), \ldots, \pi^n(s_t^n + W_n(s_t; \theta)\cdot\Delta s^n)]$. As a result, \eqref{eq:opt_adv_prob_3} can be efficiently solved by PGD. We present the pseudo-code of the attack in Alg.~\ref{alg:A3}. After $K$ steps of PGD update, we define the $i_{(n-j+1)}$ as index of the $j$-th largest value within $W(s_t;\theta_K) \in \mathbb{R}^{n}$, i.e. $W_{i_{(n)}}(s_t;\theta_K) \ge W_{i_{(n-1)}}(s_t;\theta_K) \ge\cdots\ge W_{i_{(1)}}(s_t;\theta_K)$. Let $\mathcal{I}_j$ be the index set of top-$j$ largest outputs of the $W(s_t;\theta_K)$ network. The final perturbation returned by our victim agent selection strategy will be $\widehat{\Delta s} = ((\widehat{\Delta s})^1,\cdots, (\widehat{\Delta s})^n)$ where $(\widehat{\Delta s})^i = \textbf{0}$ if $i\notin \mathcal{I}_{n_v}$ and $(\widehat{\Delta s})^i = (\Delta s_{K})^i$ if $i\in \mathcal{I}_{n_v}$.

\begin{algorithm}[hpt!]\caption{cMBA with victim agent selection at time-step $t$}\label{alg:A3}
\normalsize
\begin{algorithmic}[1]
   \STATE\label{step:i0}{\bfseries Initialization:} Given $s_t$, $s_{fail}$, $\pi$, $f$, $n_v$; initialize $\Delta s_0$; choose learning rate $\eta, \lambda > 0$.
   \STATE\hspace{0ex}\label{step:o1}{\bfseries For $k = 0,\cdots, K-1$ do}
   \vspace{0.25ex}   
  \STATE\hspace{2ex}\label{step:o2} Compute $a_t = \pi(\{s_t^i + W_i(s_t;\theta)\cdot\Delta s^i\}_i)$.
  \STATE\hspace{2ex}\label{step:o2} Compute $\hat{s}_{t+1} = f(s_t,a_t)$.
  \STATE\hspace{2ex}\label{step:o2} Update $\Delta s$ as $\Delta s_{k+1} = \proj_{\mathcal{C}_{p,\varepsilon,t}}\left[\Delta s_k - \eta \nabla_{\Delta s}d(\hat{s}_{t+1}, s_{fail})\right]$.
\STATE\hspace{2ex}\label{step:o2} Update $\theta$ as $\theta_{k+1} = \theta_k - \lambda \nabla_{\theta}d(\hat{s}_{t+1}, s_{fail})$.
   \STATE\hspace{0ex}{\bfseries End For}
    \STATE\hspace{0ex}Compute $\mathcal{I}_{n_v} = \{i_{(n)},\cdots,i_{(n-n_v)}\}$ such that $W_{i_{(n)}}(s_t,\theta_K) \ge \cdots\ge W_{i_{(1)}}(s_t,\theta_K)$.
    \STATE\hspace{0ex}Return $\widehat{\Delta s} = ((\widehat{\Delta s})^1,\cdots, (\widehat{\Delta s})^n)$ where $(\widehat{\Delta s})^i = (\Delta s_{K})^i$ if $i\in \mathcal{I}_{n_v}$ and $(\widehat{\Delta s})^i= 0$ otherwise.
\end{algorithmic}
\end{algorithm}

\begin{remark}
For this attack, we assume each agent $i$ has access to the other agent's state to form the joint state $s_t$. If the set of victim agents is pre-selected, we do not need this assumption and the adversarial attack can be performed at each agent independently of others.
\end{remark}

\begin{remark}
We can also use Alg.~\ref{alg:A3} in conjunction with Alg.~\ref{alg:A1} to perform the attack using Alg.~\ref{alg:A1} on the set of victim agents returned by Alg.~\ref{alg:A3}. Our results for Experiment~(IV) in Section~\ref{sec:experiment} indicate that using Alg.~\ref{alg:A3} and Alg.~\ref{alg:A1} together produces an even stronger cMBA attack: e.g. it can further decrease the team reward up to 267\% \textit{more} than using only Alg.~\ref{alg:A3} in \textbf{Ant (4x2)} environment.
\end{remark}

We highlight that combining cMBA with victim agent selection strategy can potentially make cMBA much stronger as shown in our comparison with other selection strategy such as random selection or greedy selection in Section~\ref{sec:experiment}. Our numerical results show that selecting the appropriate set of victim agents at each time-step constantly outperform cMBA attack with random/greedy agent selection. For example, with victim agent selection strategy, cMBA can further lower the team reward up to 267\%, 486\%, 30\% \textit{more} than when using cMBA on randomly selected, greedily selected, or fixed agent in \textbf{Ant (4x2)} environment.



\section{Experiments}\label{sec:experiment}

We perform the attack on four multi-agent MuJoCo (MA-MuJoCo) environments \citep{mamujoco} including \textbf{Ant(4x2)}, \textbf{HalfCheetah(2x3)}, \textbf{HalfCheetah(6x1)}, and \textbf{Walker2d(2x3)}. 
The pair \textbf{name(config)} indicates the name of MuJoCo environment along with the agent partition, where a configuration of 2x3 means there are in total 2 agents and each agent has 3 actions. Each of the original MuJoCo agent in the single-agent setting contains multiple joints and the way these joints are partitioned will lead to different multi-agent configurations. These configurations are described as follows:
\begin{itemize}
\setlength{\itemindent}{-.1in}
    \item \textbf{Walker (2x3)} environment: this environment has 6 joints, 3 for each leg and the whole agent is divided into 2 group of joints $\{1,2,3\}$ and $\{4,5,6\}$ representing two legs \citep[Fig. 4F]{peng2020facmac}.
    
    \item \textbf{HalfCheetah(2x3)} environment: there are two agents, each represents a front or rear leg with joints $\{1,2,3\}$ and $\{4,5,6\}$ \citep[Fig. 4C]{peng2020facmac}.
    
    \item \textbf{HalfCheetah(6x1)} environment: each agent represents each of the total 6 joints \citep[Fig. 4D]{peng2020facmac}.
    
    \item \textbf{Ant(4x2)} environment: each agent controls one leg with two joints out of 4 legs \citep[Fig. 4J]{peng2020facmac}.
\end{itemize}
We also demonstrate the effectiveness of the learned failure state approach using the multi-agent particle environment, denoted as \textbf{MPE(3x5)} \citep{lowe2017multi, mordatch2017emergence} where we do not have expert knowledge of the failure state. For all environments, we use MADDPG \citep{lowe2017multi} to obtain trained MARL agents. To perform the attack, we consider the following \textbf{model-free} baselines:
\begin{enumerate}\setlength\itemsep{-0.075ex}
    \item \textbf{Uniform}: the perturbation follows the Uniform distribution $U(-\varepsilon,\varepsilon)$.
    \item \textbf{Gaussian}: the perturbation follows the Normal distribution $\mathcal{N}(0,\varepsilon)$.
    \item \textbf{Lin et al. (2020) + iFGSM}: Since there is no other work performing adversarial attack for continuous action space in c-MARL, we adapt the approach in \citep{lin2020robustness} to form another baseline. 
    In particular, we train an adversarial policy for one agent to minimize the total team reward while the remaining agents use the trained MARL policy. This adversarial policy is trained for 1 million timesteps. We then use this trained policy to generate a "target" action and use iterative FGSM method \citep{kurakin2016adversarial,goodfellow2014explaining} to generate the adversarial observation perturbation for the agents' input. Note that the adversarial policy is trained on the same agent that is being attacked.
    We consider two variants of this approach. The first variant is denoted as \textbf{Lin et al (2020) + iFGSM} where the attack for any agent uses a shared adversarial policy trained for agent $0$. The second variant is denoted as \textbf{Lin et al (2020) + iFGSM (*)} where the adversarial policy is trained on the same agent when performing the attack on that agent.
\end{enumerate}

In our experiments, we consider two variants of c-MBA:
\begin{enumerate}
    \item \textbf{c-MBA-F}: we perform c-MBA using an expert-defined failure state (see below) where we have the knowledge of the state definition of the c-MARL environment. 
    \item \textbf{c-MBA-D}: we perform c-MBA attack using failure state learned from the collected data as described in Section~\ref{subsec:data_drive_failure_state}.
\end{enumerate}

\textbf{Specifying target observation for each environment:} To perform our model based attack, we need to specify a target observation that potentially worsens the total reward. Currently, we do not have a general procedure to specify this target observation. We specify the target observations based on prior knowledge about the environments as follows. In multi-agent MuJoCo environments, each agent has access to its own observation of the agent consisting the position-related and velocity-related information. The position-related information includes part of $x,y,z$ coordinates and the quarternion that represents the orientation of the agent. The velocity-related information contains global linear velocities and angular velocities for each joint in a MuJoCo agent. We refer the reader to \citep{todorov2012mujoco} for more information about each MuJoCo environment. Now we describe the design of this target observation for each environment as follows:
\begin{itemize}
    \item \textbf{Walker(2x3)} environment: Since the episode ends whenever the agent falls, i.e. the $z$ coordinate falls below certain threshold. In this environment, the target observation has a value of 0 for the index that corresponds to the $z$ coordinate of the MuJoCo agent (index 0).
    
    \item \textbf{HalfCheetah(2x3)} and \textbf{HalfCheetah(6x1)} environments: As the goal is to make agent moves as fast as possible, we set the value at index corresponding to the linear velocity to 0 (index 8).
    
    \item \textbf{Ant(4x2)} environment: As the agent can move freely in a 2D-plan, we set the index corresponding to the $x,y$ linear velocities to 0 (indices 13 and 14).
\end{itemize}

\textbf{Learning the dynamics model:} For each environment, we collect 1 million transitions using the trained MARL policy $\pi_{tr}$ and a random policy $\pi_{rd}$. We partition the collected data into train and test set with a ratio 90-10. The dynamics model is represented by a fully-connected neural network. The network contains 4 hidden layers with 1000 neurons at each layer and the activation function for each hidden layer is ReLU. We train the network for 100 epochs using AdamW \citep{loshchilov2017decoupled} with early stopping where the learning rate is tuned in the set $\{0.001,0.0005,0.0001,0.00005, 0.00001\}$ to obtain the model with the best test mean squared error. Please refer to our submitted code for further details.

\textbf{Computing resources:} All experiments are implemented in Python (the code will be made publicly available after the paper gets through review process.) running on a computer with configuration: 3.2GHz 16-core Intel processors, 32GB RAM, and NVIDIA 3080ti GPU.

We evaluate c-MBA comprehensively with the following 6 experiments:
\begin{itemize}\setlength\itemsep{-0.075ex}
    \item \textbf{Experiment (I) -- model-free baselines vs model-based attack c-MBA using $\ell_{\infty}$-constrained perturbation}: we compare c-MBA-F and c-MBA-D with other baselines when attacking individual agent under $\ell_\infty$ constraint.
    \item \textbf{Experiment (II) -- effectiveness of learned adaptive victim selection}: we evaluate the performance of Alg.~\ref{alg:A3} with other heuristicly selected variants of c-MBA. 
    Our results show that in \textbf{HalfCheetah(6x1)} environment, under $\varepsilon=0.05$ and $\ell_\infty$ budget constraint, the amount of team reward reduction of our learned agents selection scheme can be up to 80\% \textit{more} than the cases when using heuristic strategy to select victim agent.
    \item \textbf{Experiment (III) -- attacking multiple agents using model-free baselines vs model-based attack c-MBA with $\ell_{\infty}$ perturbation}: we report results on attacking multiple agents simultaneously. 
    This setting is not considered in \citep{lin2020robustness}.
    \item \textbf{Experiment (IV) -- model-free baselines vs model-based attack c-MBA in MPE(3x5) environment}: we compare c-MBA-D with other baselines under $\ell_\infty$ and $\ell_1$ constraint.
    \item \textbf{Experiment (V) -- model-free baselines vs model-based attack c-MBA on $\ell_{1}$ perturbation}: we compare c-MBA-F and c-MBA-D with other baselines when attacking individual agent under $\ell_1$ constraint. 
    \item \textbf{Experiment (VI) -- adversarial attacks using dynamics model with various accuracy}: we illustrate the performance of c-MBA when using dynamics model with various accuracy. 
\end{itemize}

\begin{figure}[ht!]
    \begin{center}
    \includegraphics[width=0.23\linewidth,valign=t]{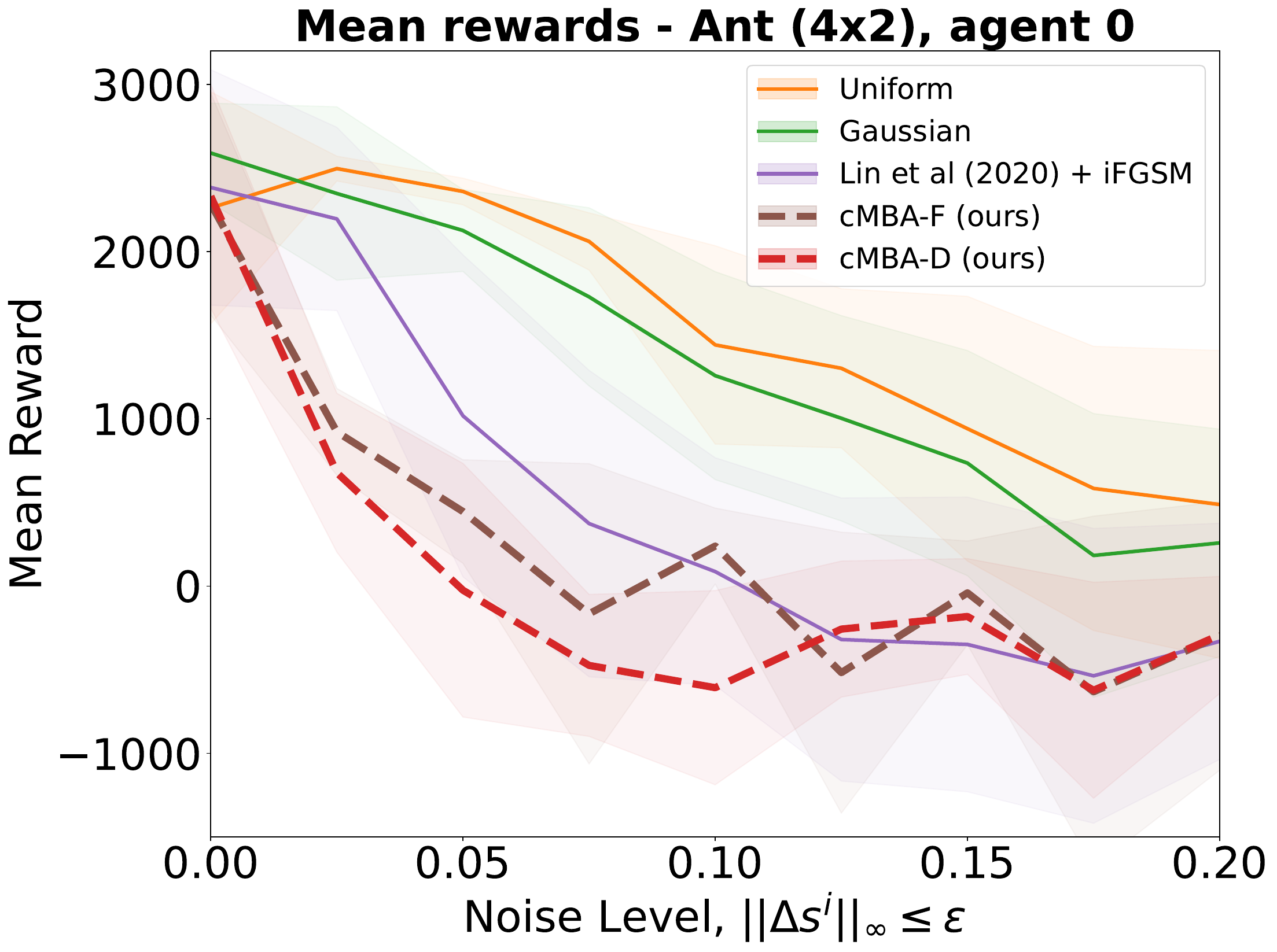}
    \includegraphics[width=0.23\linewidth,valign=t]{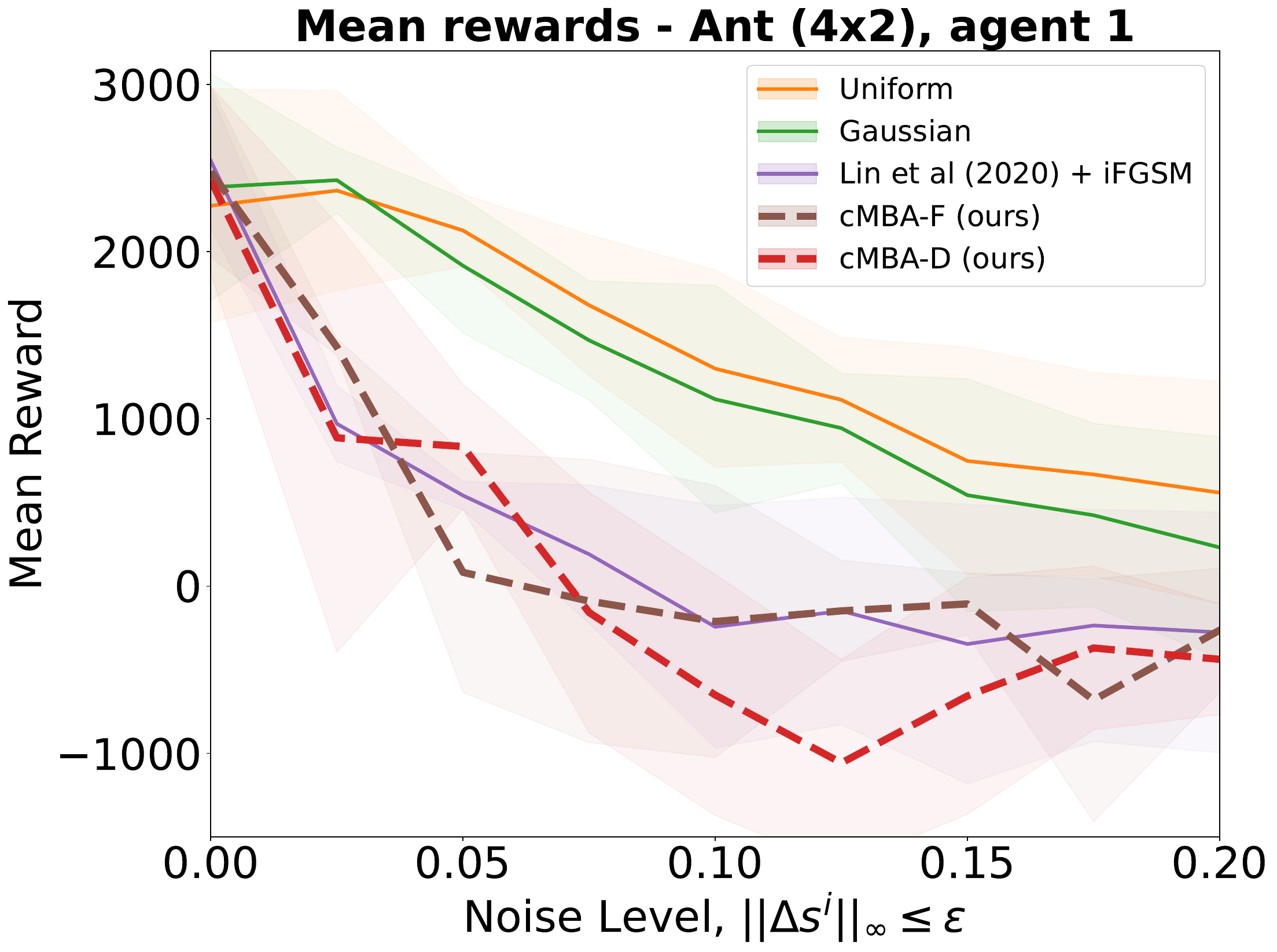}
    \includegraphics[width=0.23\linewidth,valign=t]{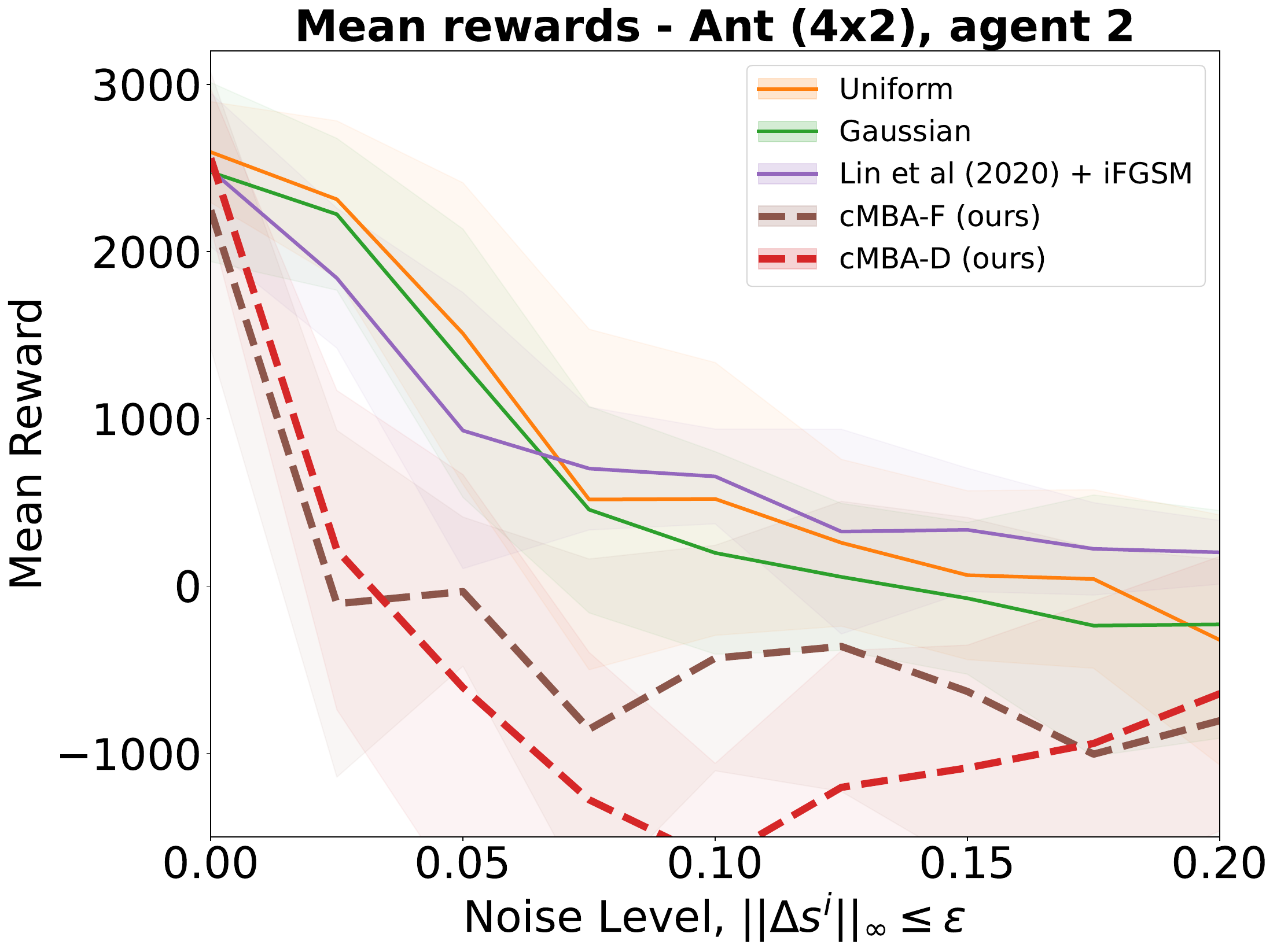}
    \includegraphics[width=0.23\linewidth,valign=t]{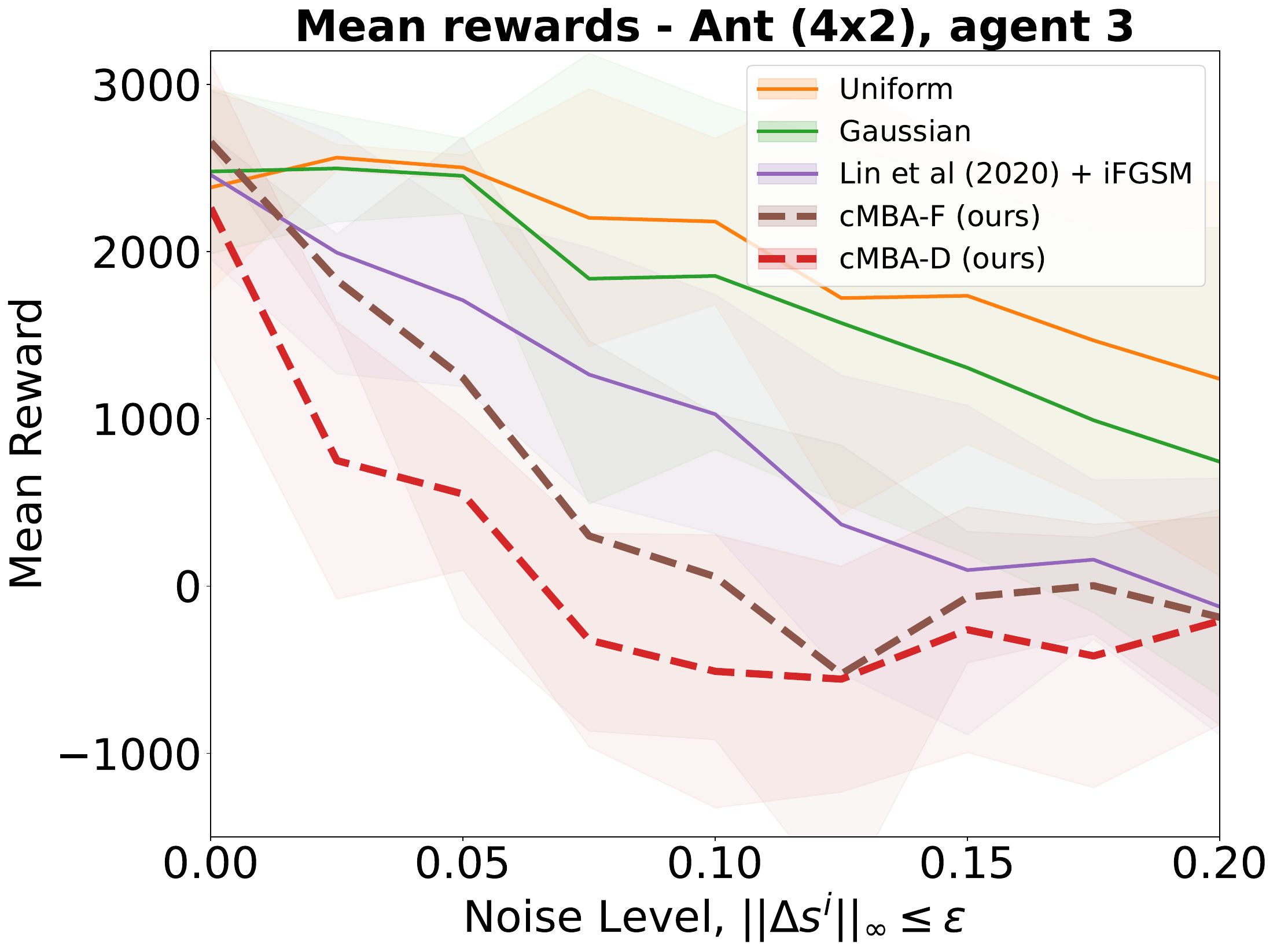}
    \includegraphics[width=0.23\linewidth,valign=t]{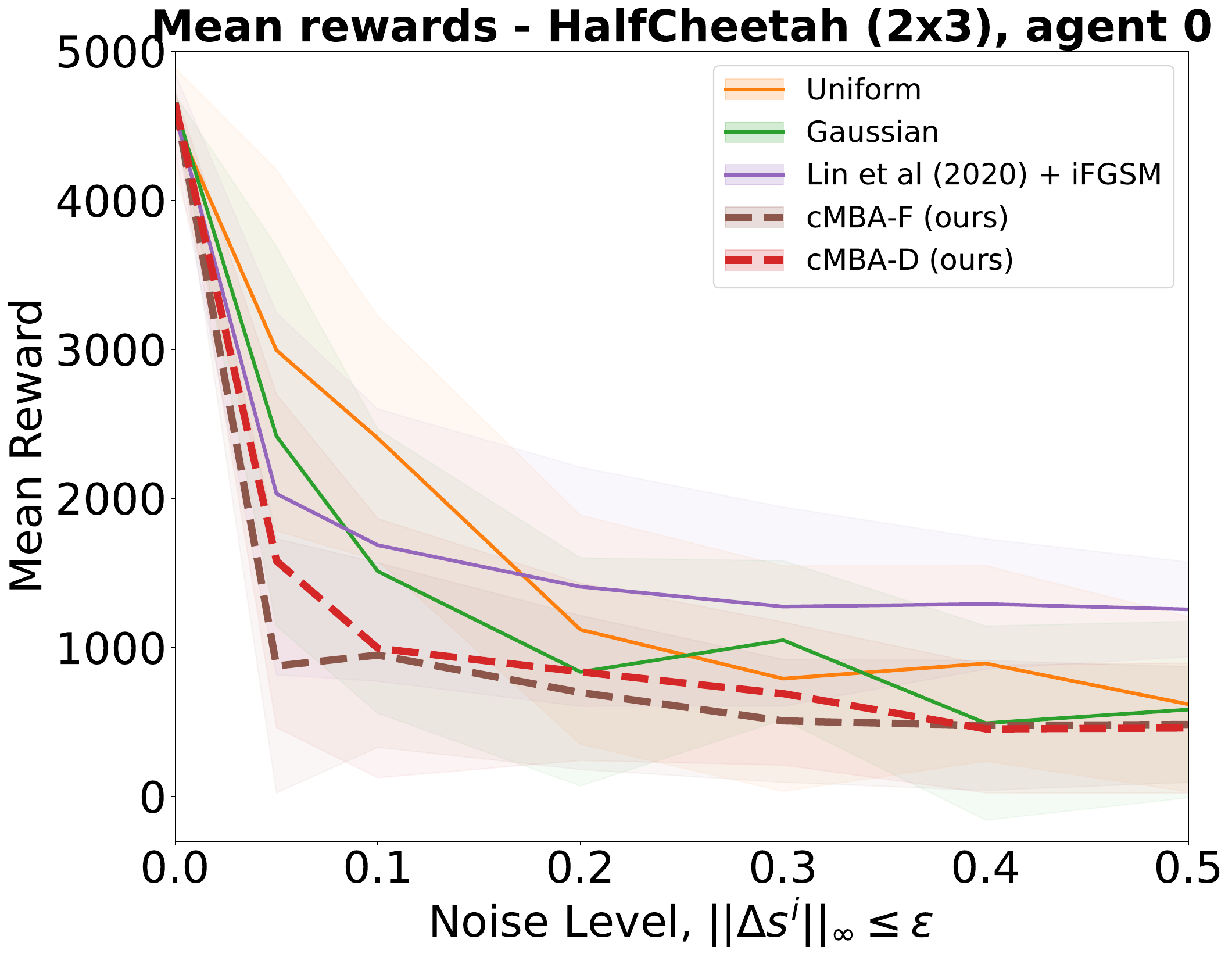}
    \includegraphics[width=0.23\linewidth,valign=t]{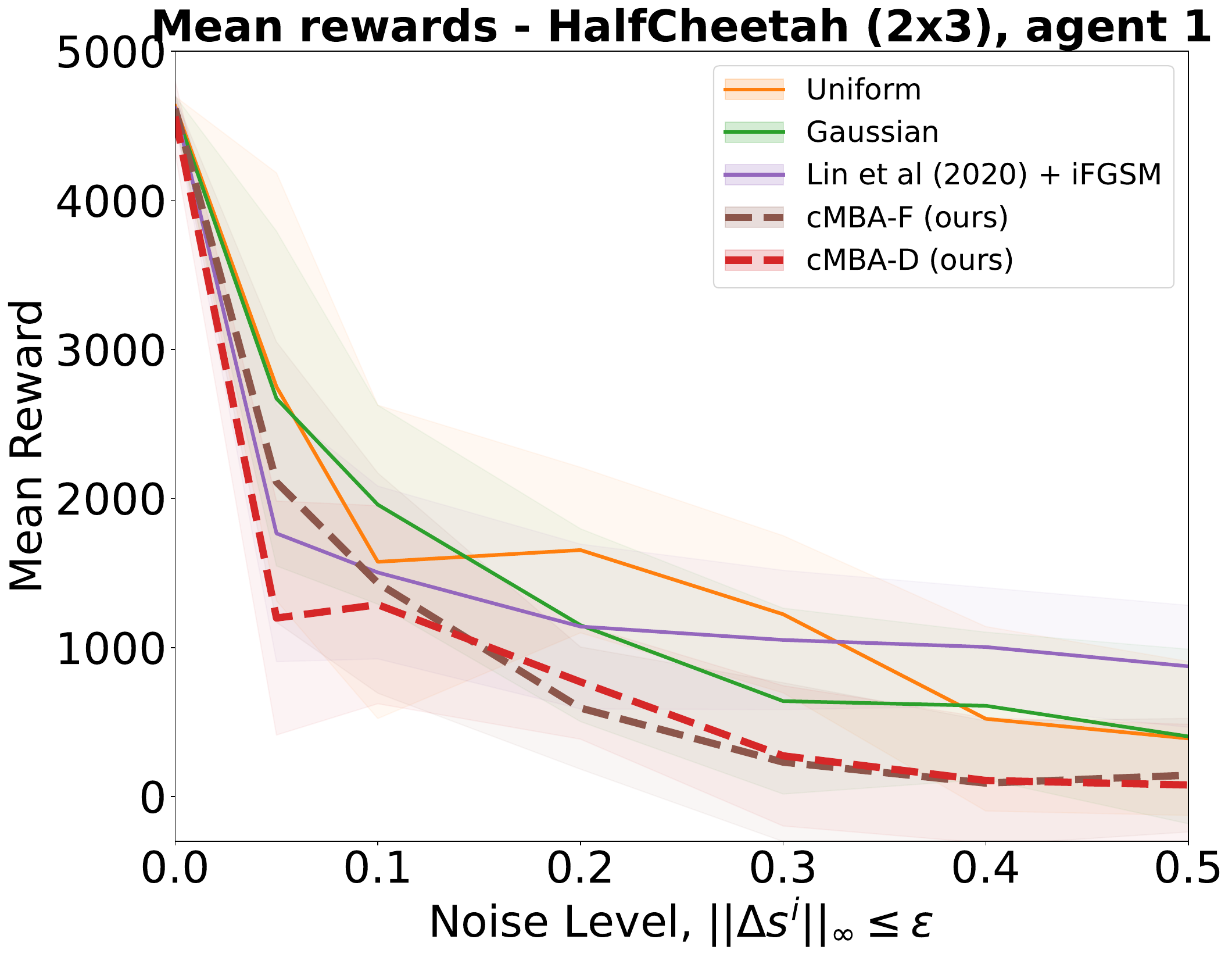}
    \includegraphics[width=0.23\linewidth,valign=t]{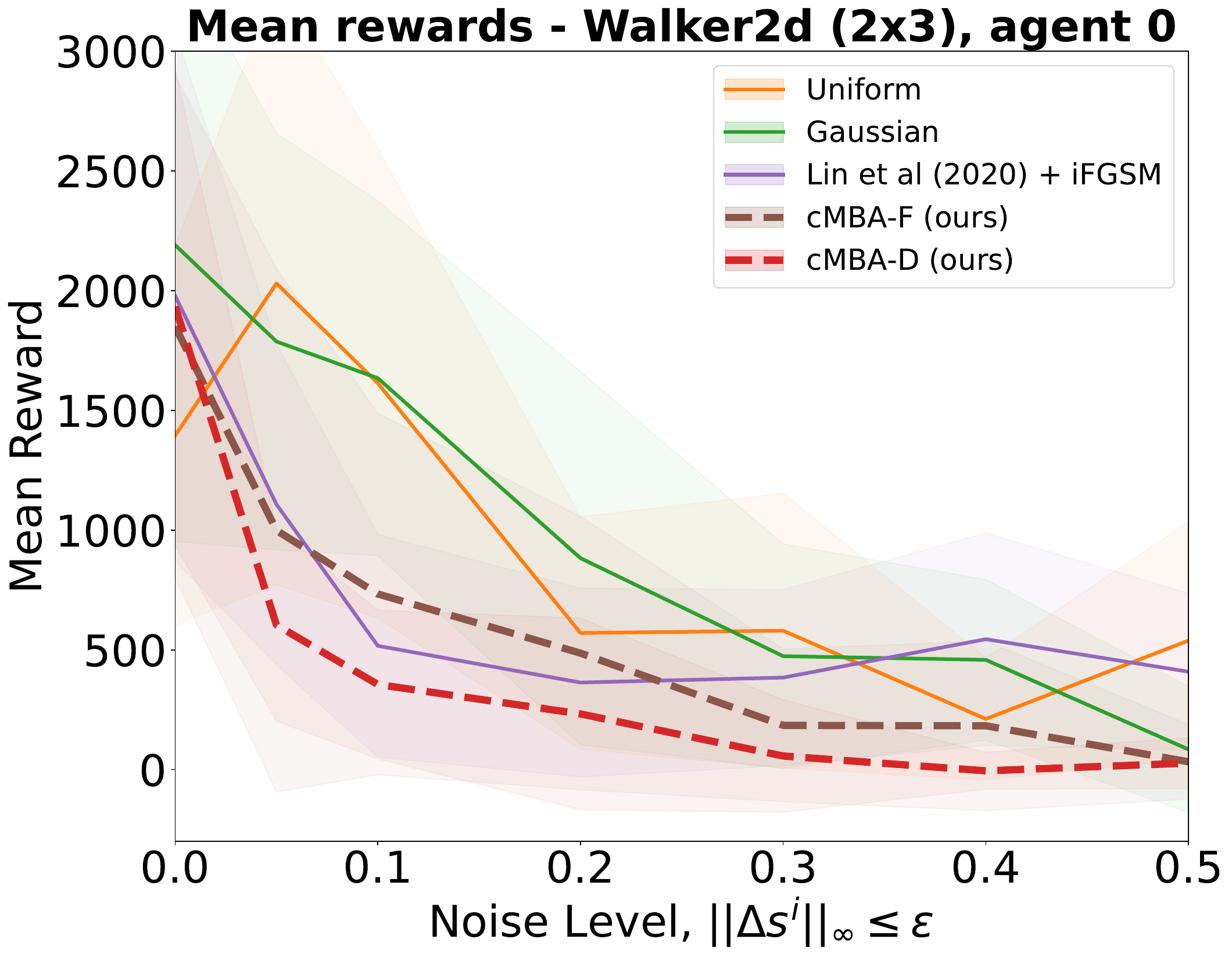}
    \includegraphics[width=0.23\linewidth,valign=t]{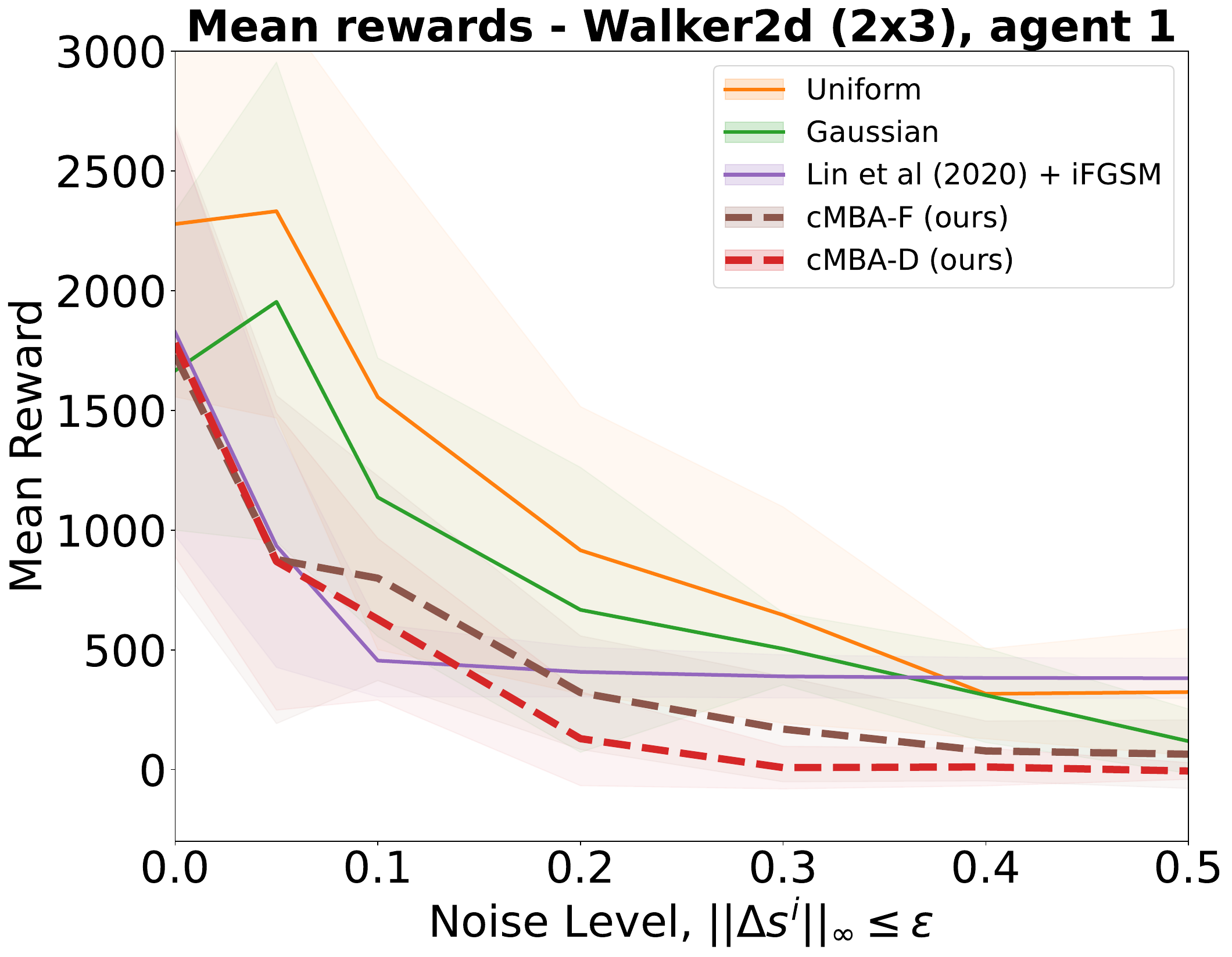}
    \caption{c-MBA vs baselines when attacking one agent in \textbf{Ant(4x2)}, \textbf{HalfCheetah(2x3)}, and \textbf{Walker2d(2x3)} environments - \textbf{Exp. (I)}.}
    \label{fig:base_vs_model_ant_cheetah2x3_linf}
    \end{center}
    \vspace{-1ex}
\end{figure}

\begin{figure}[ht!]
    \begin{center}
    \includegraphics[width=0.33\linewidth,valign=t]{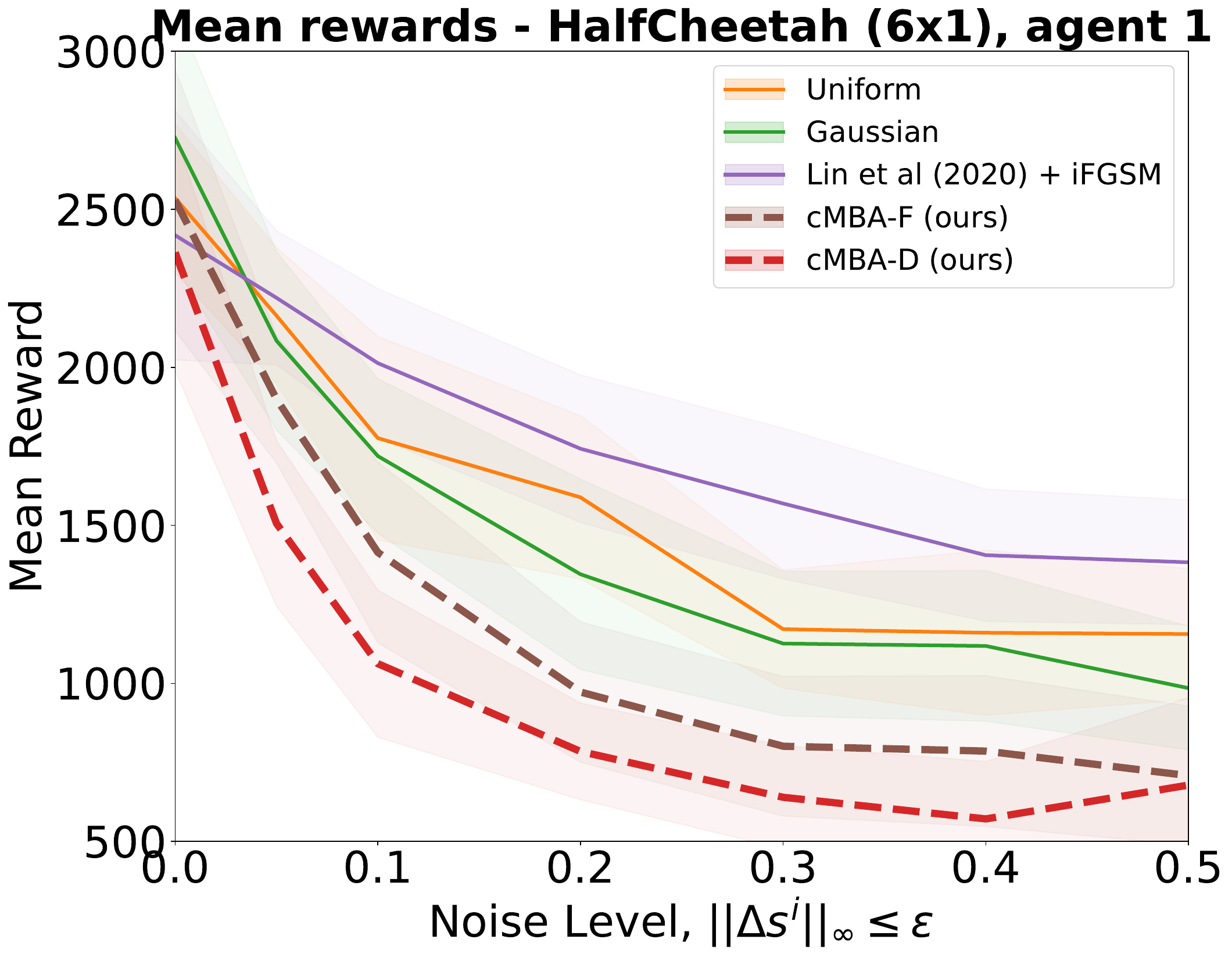}
    \includegraphics[width=0.33\linewidth,valign=t]{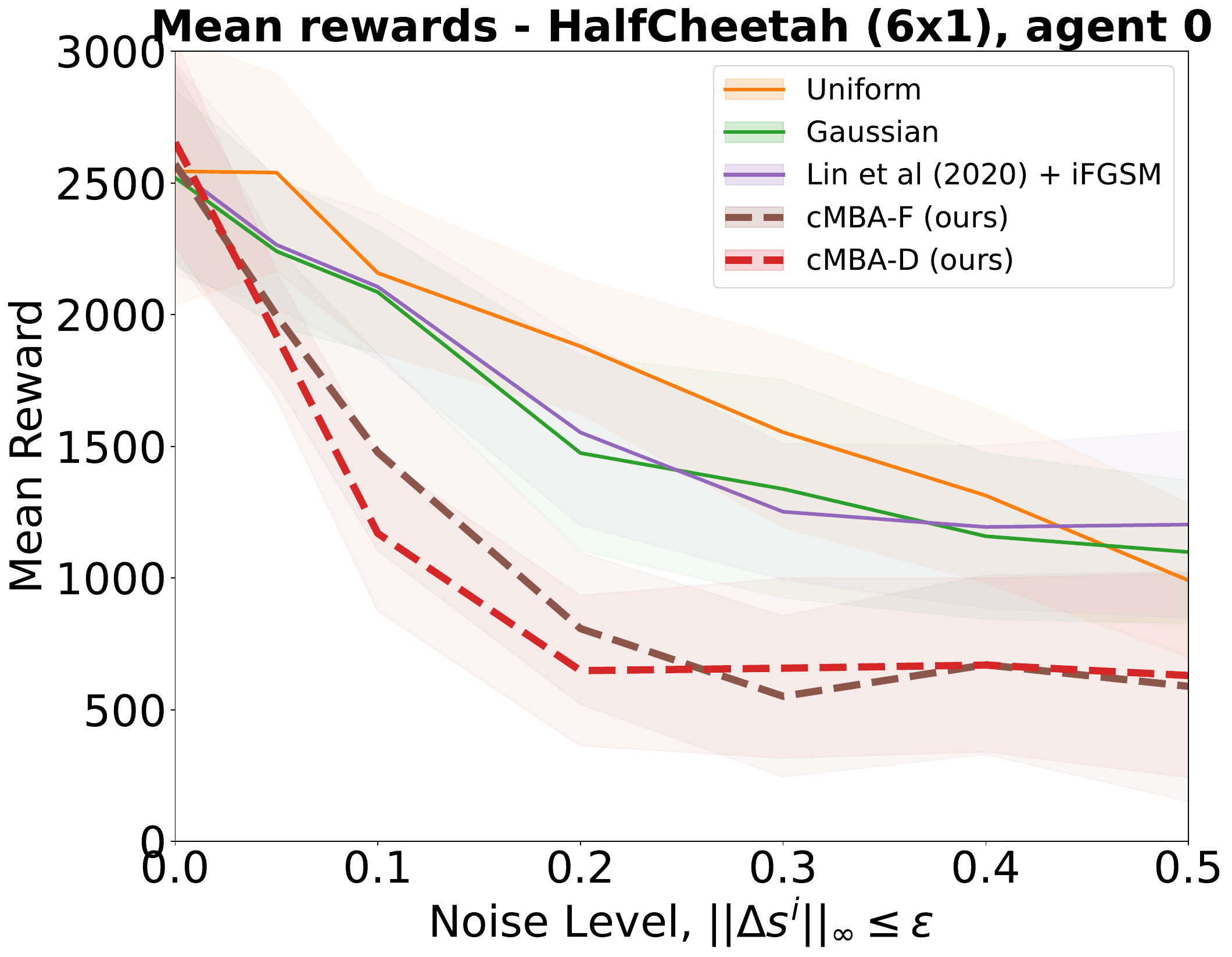}
    \includegraphics[width=0.33\linewidth,valign=t]{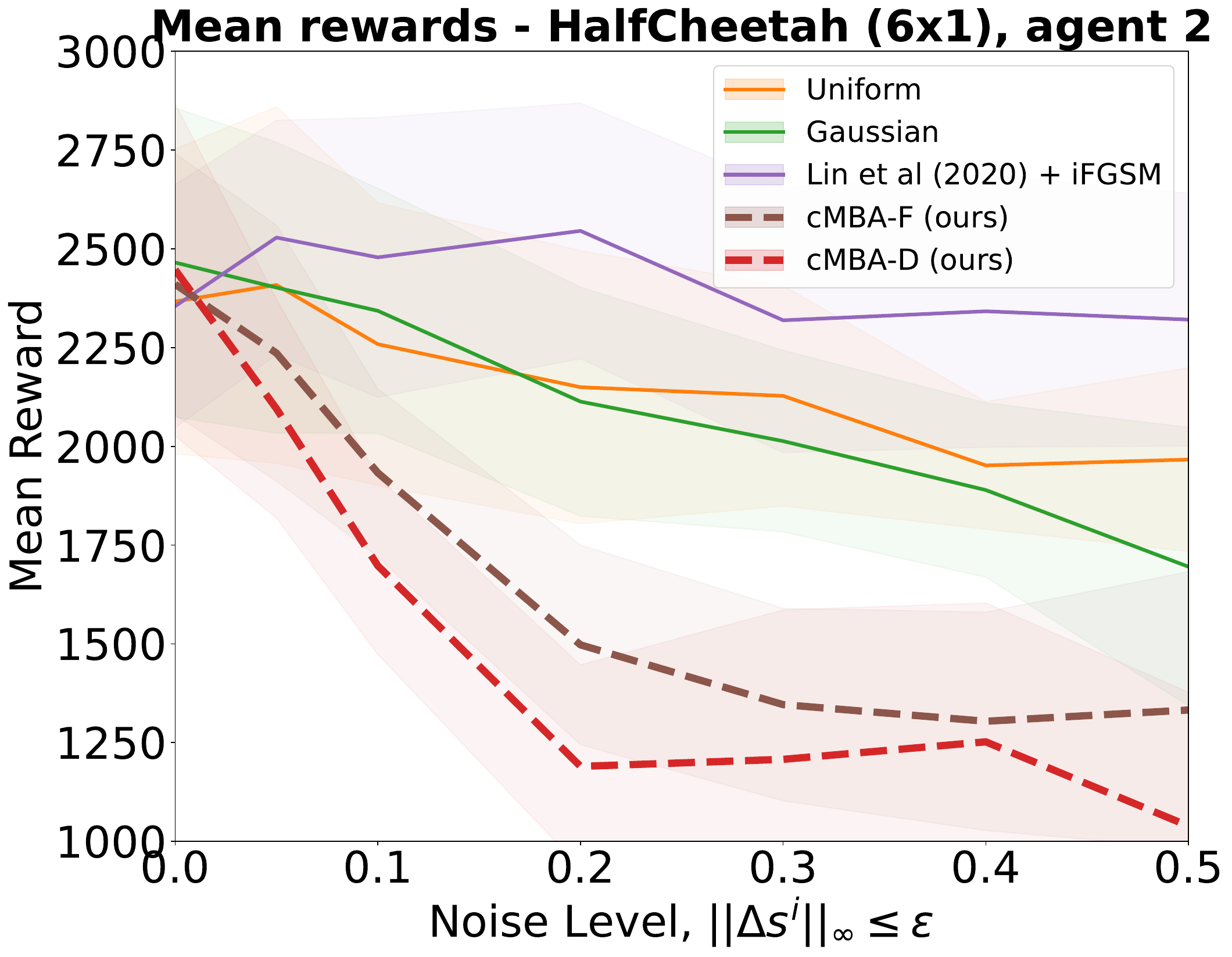}
    \includegraphics[width=0.33\linewidth,valign=t]{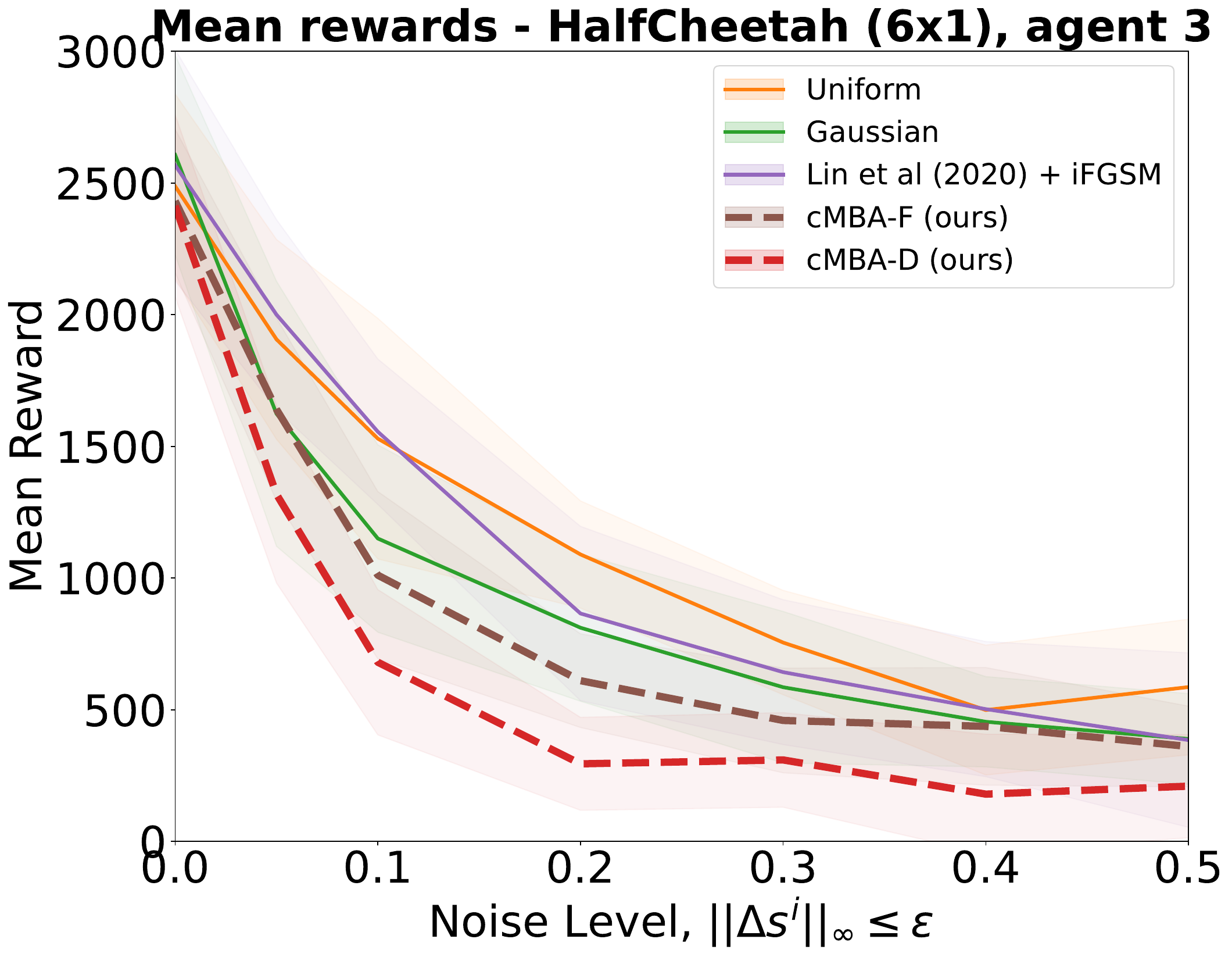}
    \includegraphics[width=0.33\linewidth,valign=t]{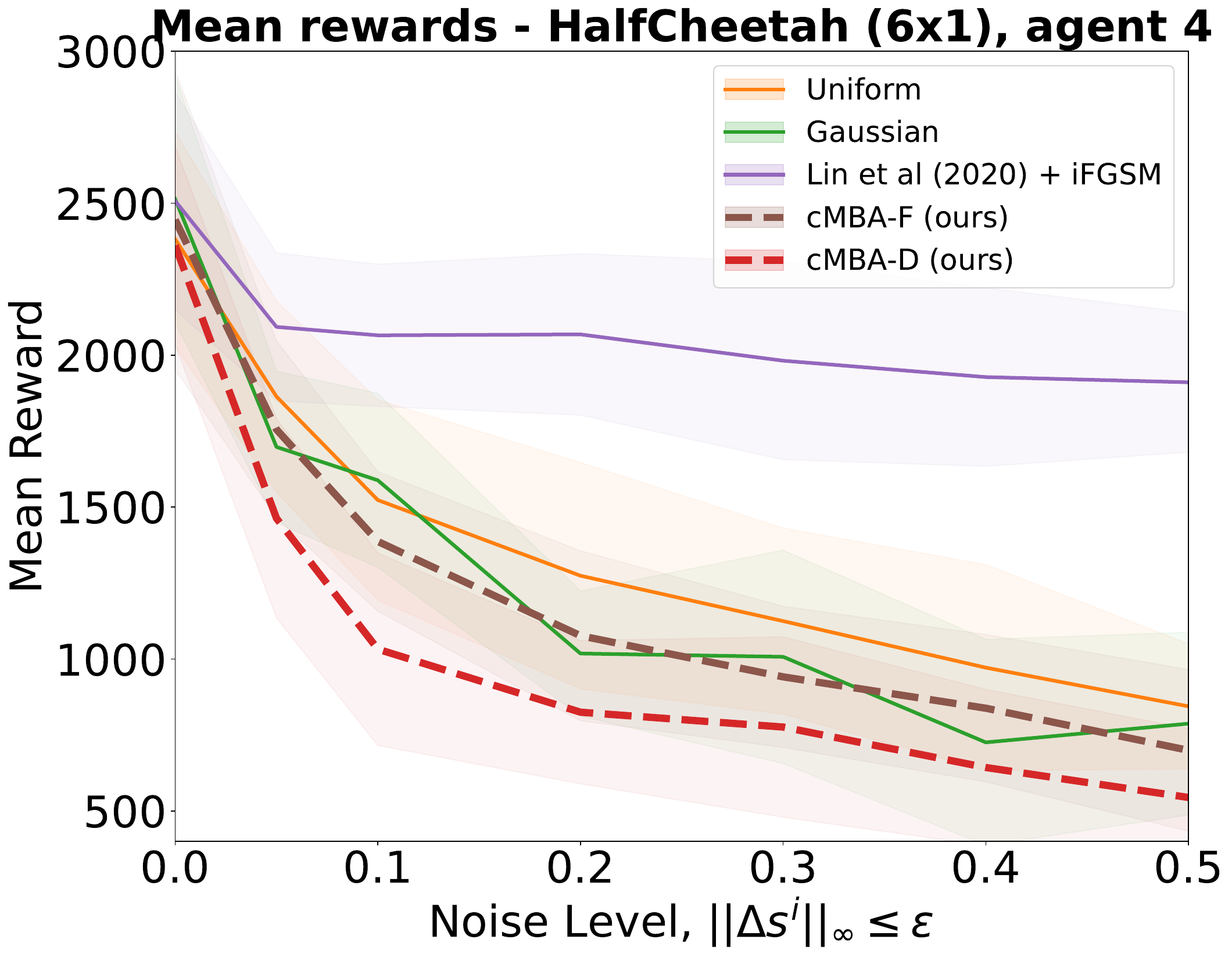}
    \includegraphics[width=0.33\linewidth,valign=t]{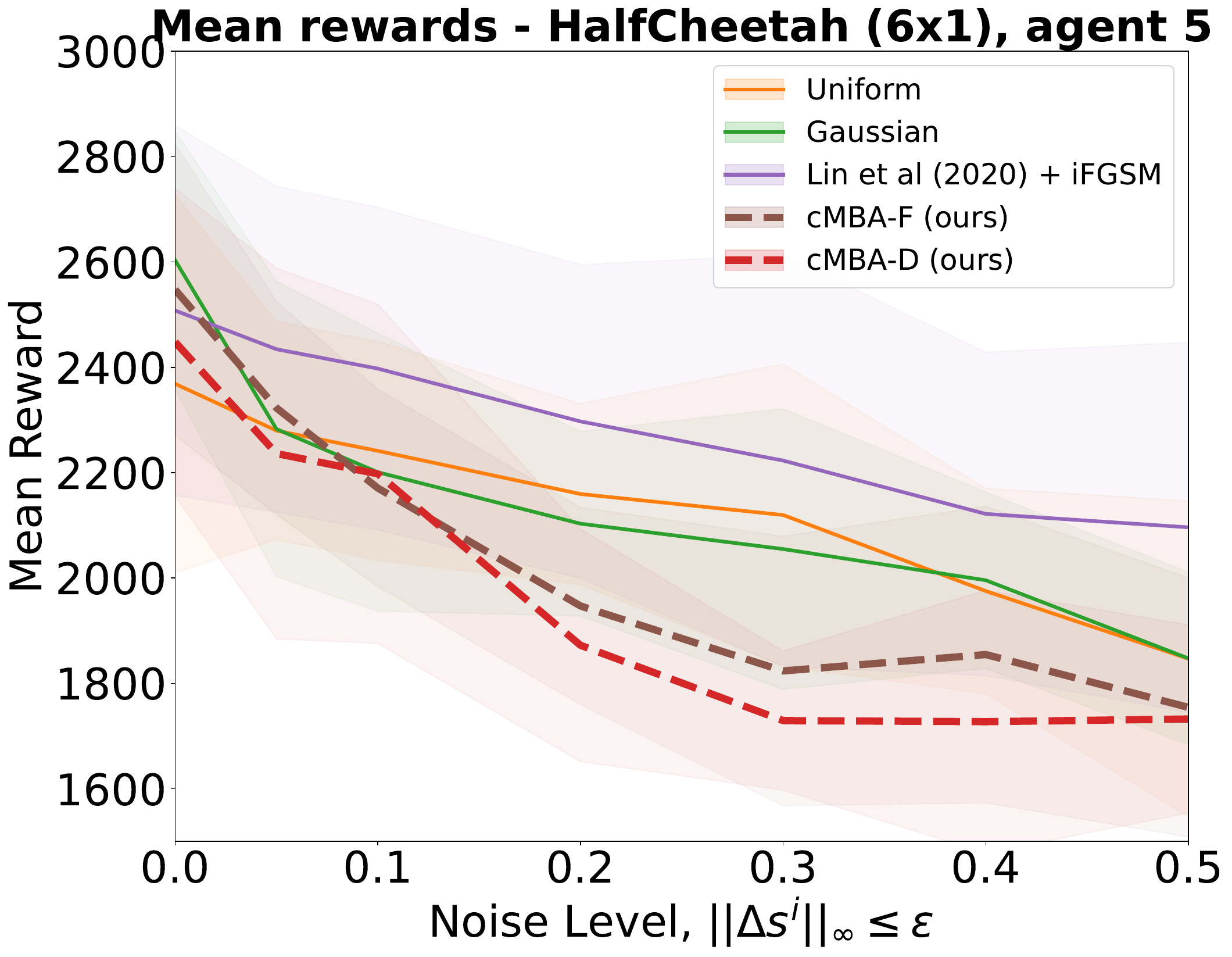}
    \caption{c-MBA vs baselines when attacking one agent in \textbf{HalfCheetah(6x1)} environment - \textbf{Exp. (I)}.}
    \label{fig:base_vs_model_ant_cheetah2x3_linf_2}
    \end{center}
    \vspace{-1ex}
\end{figure}

\textbf{Experiment (I) -- model-free baselines vs model-based attack c-MBA on $\ell_{\infty}$ perturbation.} In this experiment, we compare the 3 baseline attacks with two variants of our model-based attack on the four MA-MuJoCo environments with one victim agent ($n_v = 1$) in coherence with \citep{lin2020robustness}. 
Fig.~\ref{fig:base_vs_model_ant_cheetah2x3_linf} and Fig.~\ref{fig:base_vs_model_ant_cheetah2x3_linf_2} 
illustrate the performance when we perform these attacks on each agent with different attack budget using $\ell_\infty$-norm.
For a fixed agent, our model-based attack outperforms all the other baselines. In particular, our model-based attack yields much lower rewards under relatively low budget constraints (when $\varepsilon \in [0.05,0.2]$) compared to other baselines. For instance, under budget level $\varepsilon =0.05$, the amount of team reward reduction of our \textbf{c-MBA} attack is (on average) 66\%, 503\%, and 806\% \textit{more} than the three model-free (Lin et al. (2020) + iFGSM, Gaussian, and Uniform, respectively) in \textbf{Ant(4x2)} environment. We also observe that \textbf{c-MBA-D} either matches or performs better than \textbf{c-MBA-F} with the expert-defined failure state.

\textbf{Experiment (II) -- effectiveness of learned adaptive victim selection.} To evaluate our proposed victim-agent selection strategy, we consider the following variants of our model-based attack: 
\begin{enumerate}\setlength\itemsep{-0.05ex}
    \item \textbf{c-MBA(fixed agents)}: attack a fixed set of victim agents with Alg.~\ref{alg:A1}. We use \textbf{c-MBA (best fixed agents)} to denote the best result among fixed agents.
    \item \textbf{c-MBA(random agents)}: uniformly randomly select victim agents to attack with Alg.~\ref{alg:A1}.  
    \item \textbf{c-MBA(greedy agents selection)}: for each time step, sweep all possible subsets of agents with size $n_v$ and perform Alg.~\ref{alg:A1}. Select the final victim agents corresponding to the objective value (distance between predicted observation and target failure observation). 
    \item \textbf{c-MBA(learned agents selection)}: use Alg.~\ref{alg:A3} to attack the most vulnerable agents.
    \item \textbf{c-MBA(learned agents selection + Alg. 1)}: use Alg.~\ref{alg:A3} to select the most vulnerable agents to attack then perform the attack with the selected agents with Alg.~\ref{alg:A1}.
\vspace{-1ex}
\end{enumerate}

\begin{figure}[ht!]
    \begin{center}
    \includegraphics[width=0.24\linewidth,valign=t]{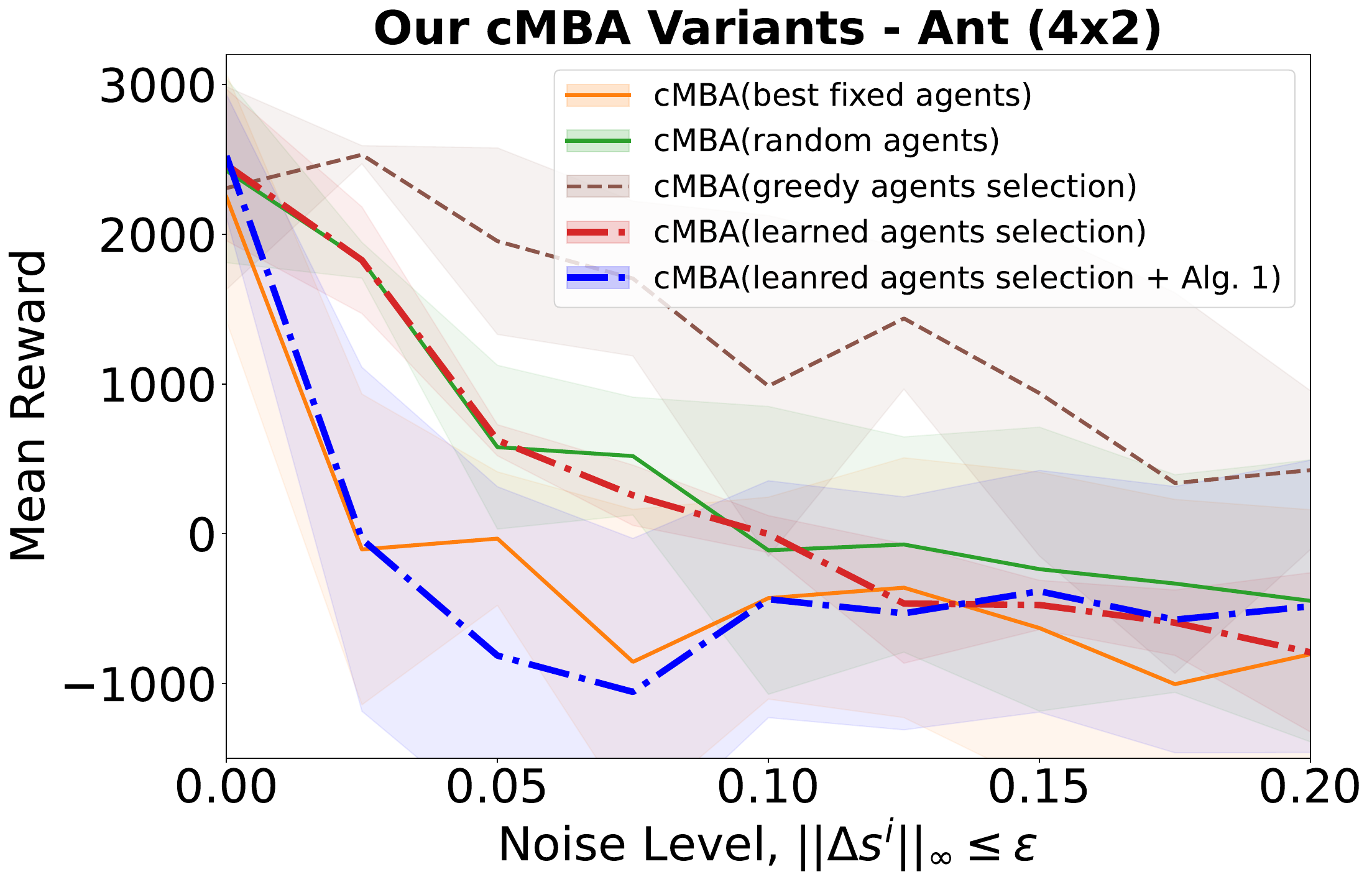}
    \includegraphics[width=0.24\linewidth,valign=t]{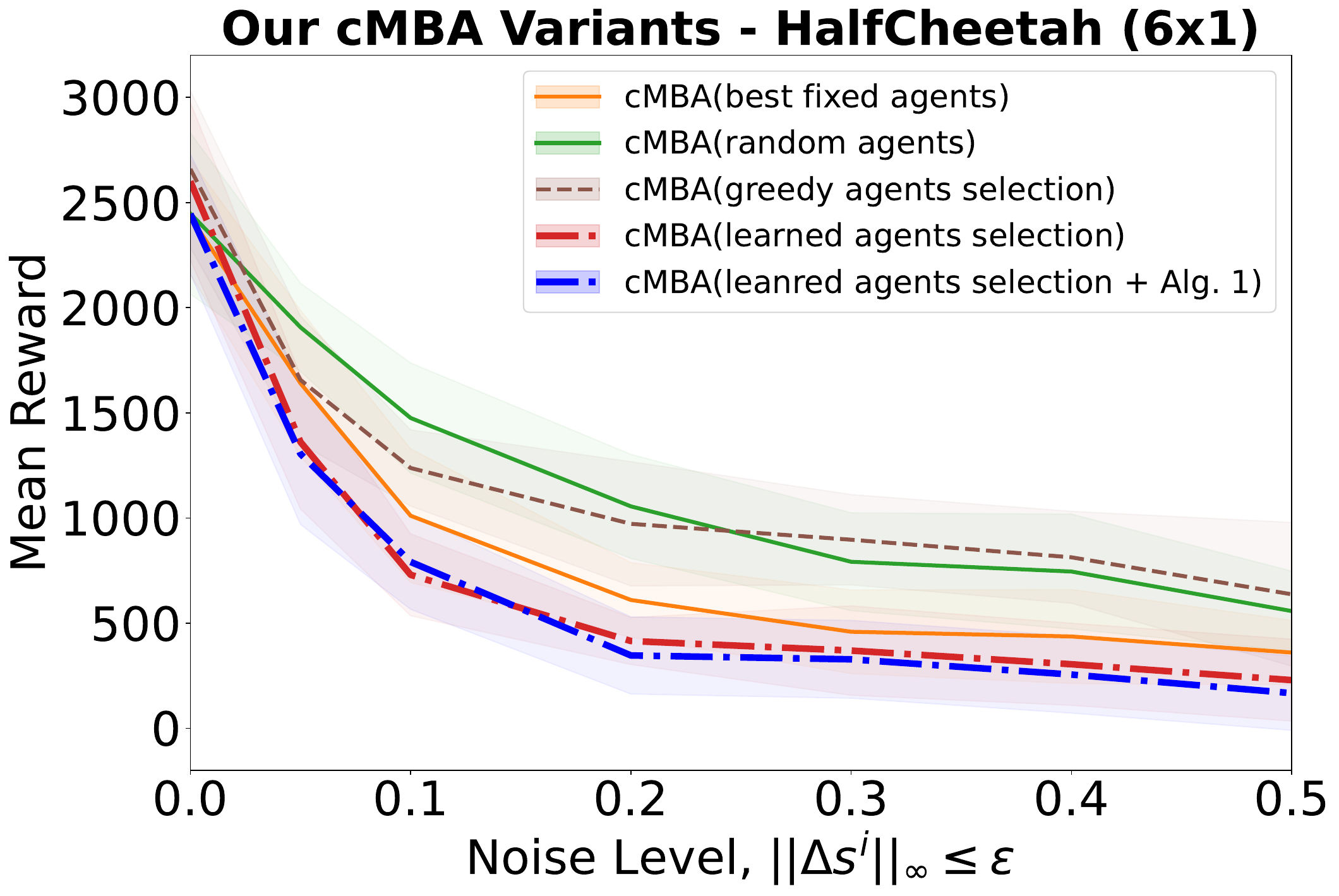}
    \includegraphics[width=0.24\linewidth,valign=t]{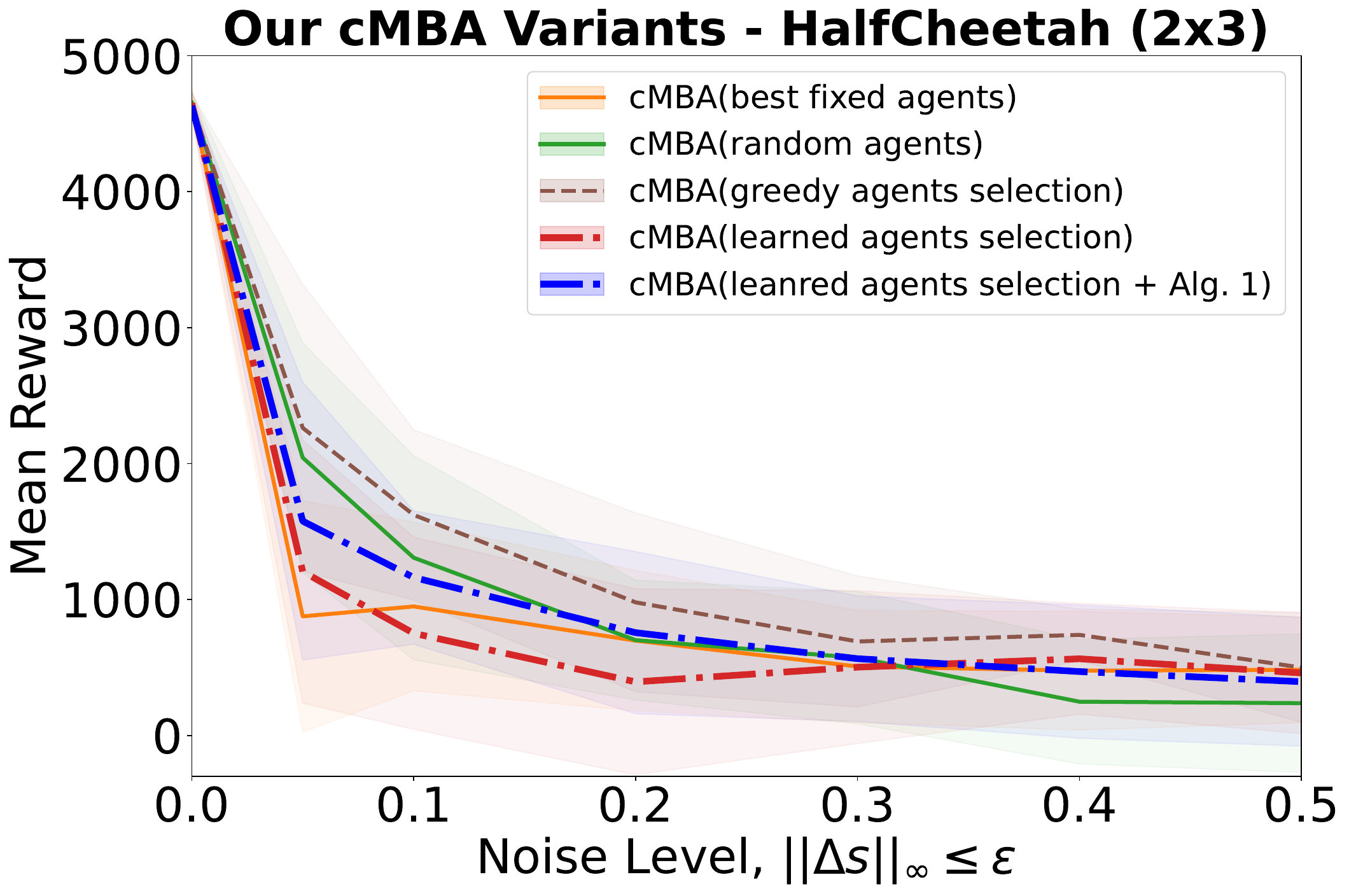}
    \includegraphics[width=0.24\linewidth,valign=t]{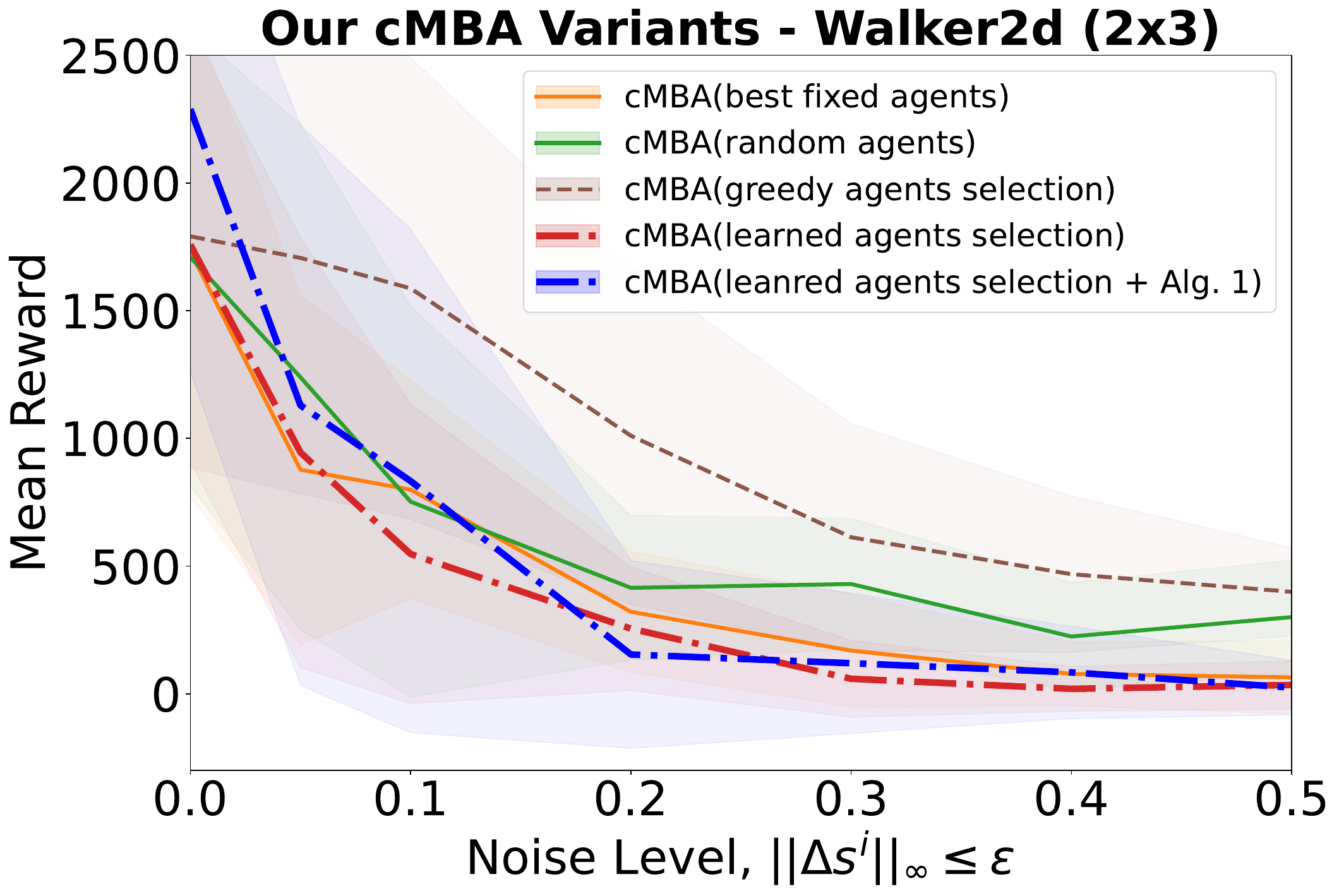}
    \caption{Performance of c-MBA with different victim agent selection strategies - \textbf{Exp. (II)}.}
    \label{fig:model_variants}
    \end{center}
    \vspace{-1ex}
\end{figure}

Fig.~\ref{fig:model_variants} illustrates the results on MA-MuJoCo environments. \textbf{c-MBA(learned agents selection)} and \textbf{c-MBA(learned agents selection + Alg. 1)} appear to be better than the random or greedy strategy and is comparable with the randomly selected one in HalfCheetah(2x3) environment. It is interesting to observe that \textbf{c-MBA(learned agents selection + Alg. 1)} is either comparable or better than \textbf{c-MBA(learned agents selection)} which shows that running Alg.~\ref{alg:A1} using agents selected by Alg.~\ref{alg:A3} can be beneficial. We show that the victim selection technique is really important as randomly choosing the agents to attack (i.e. \textbf{c-MBA(random agents)}) or even choosing agents greedily (i.e. \textbf{c-MBA(random agents)}) cannot form an effective attack and are in fact, even worse than the attacking on fixed agents (i.e \textbf{c-MBA (best fixed agents)}) in most cases. For example, our results show that in \textbf{HalfCheetah(6x1)} environment, under $\varepsilon=0.05$ and $\ell_\infty$ budget constraint, the amount of team reward reduction of our learned agents selection scheme is 33\%, 80\%, 35\% \textit{more} than the cases when attacking fixed agents, random agents, or greedily selected agents, respectively.

\begin{figure}[ht!]
    \begin{center}
    \includegraphics[width=0.33\linewidth,valign=t]{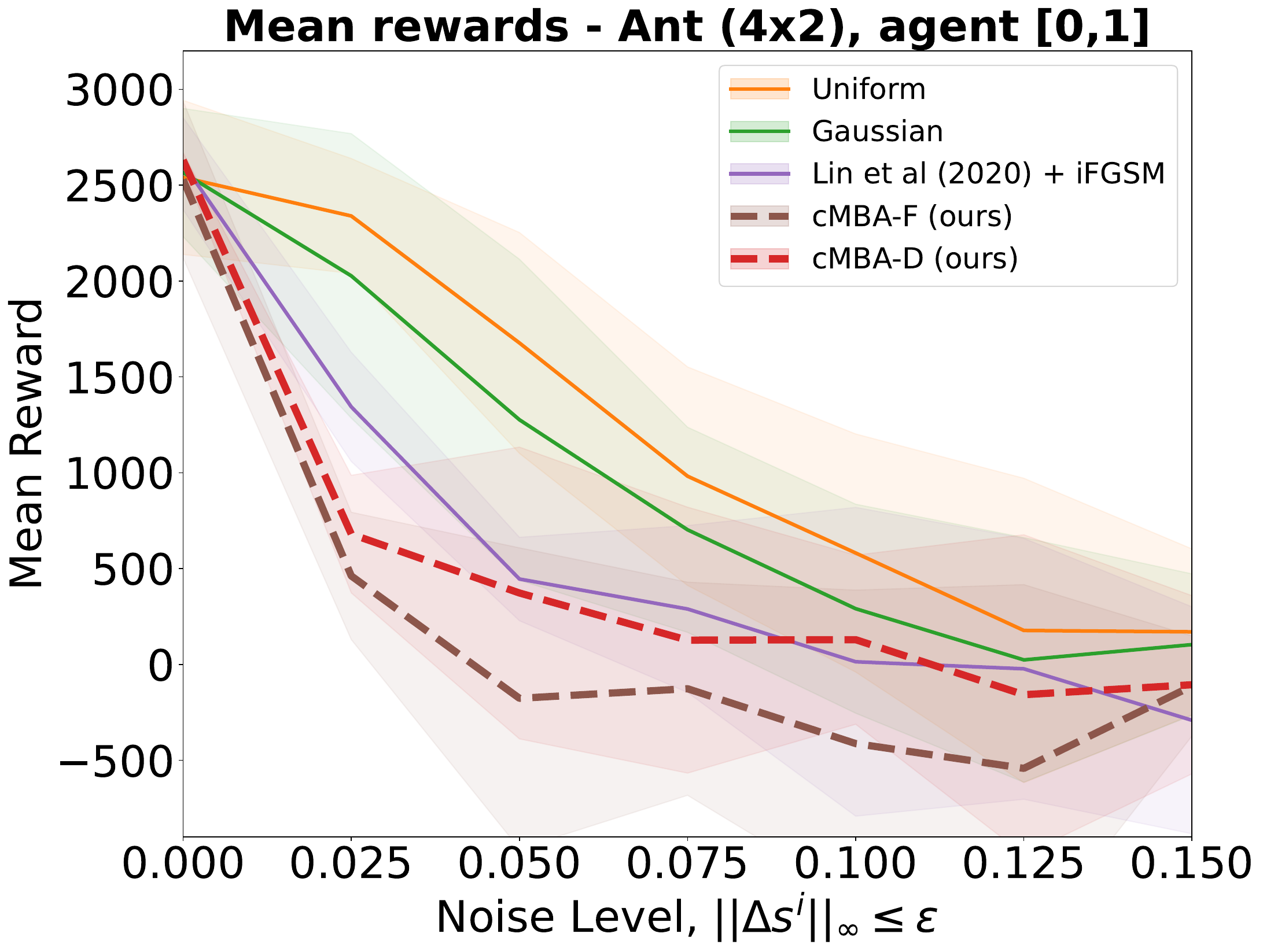}
    \includegraphics[width=0.33\linewidth,valign=t]{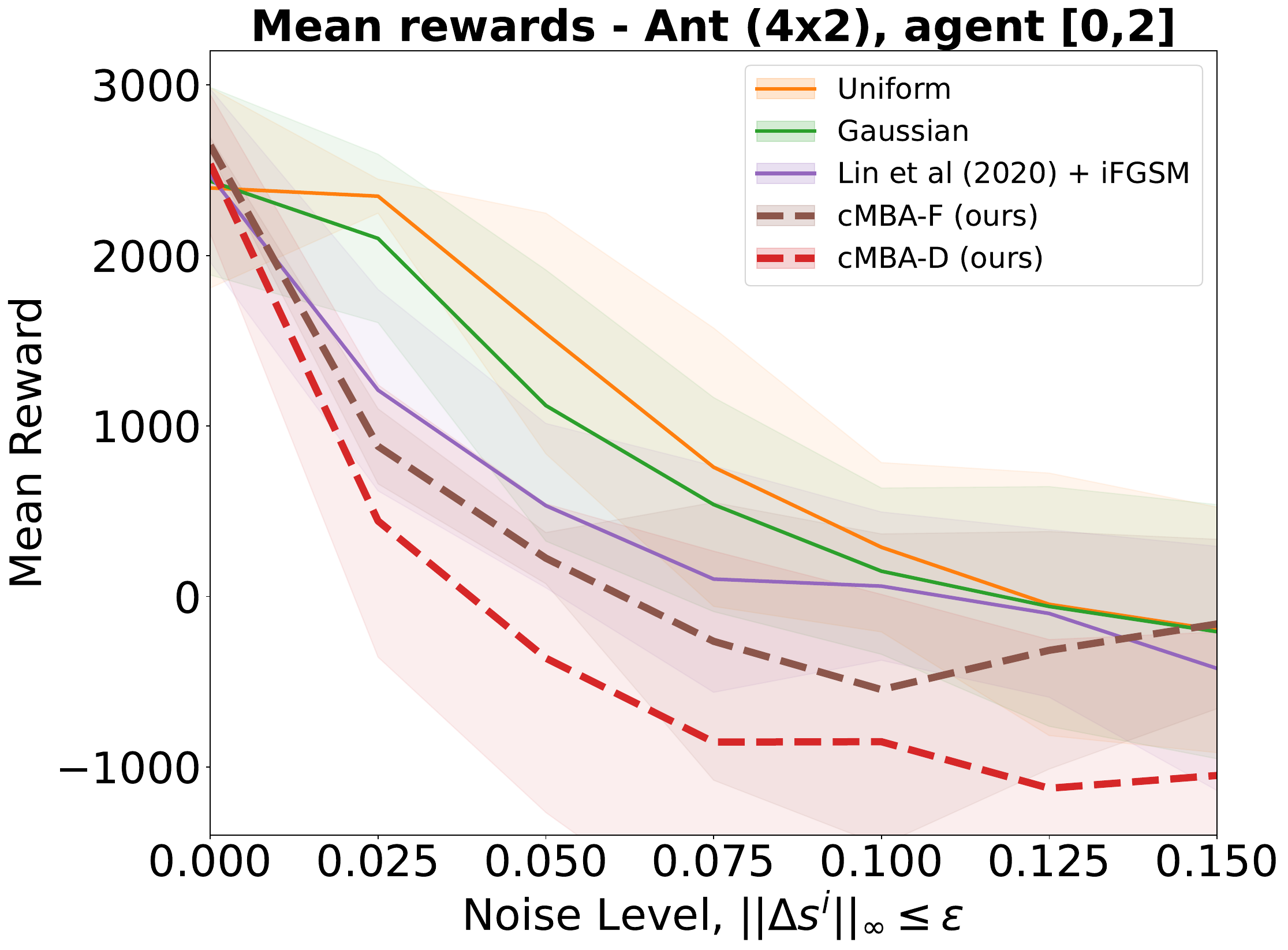}
    \includegraphics[width=0.33\linewidth,valign=t]{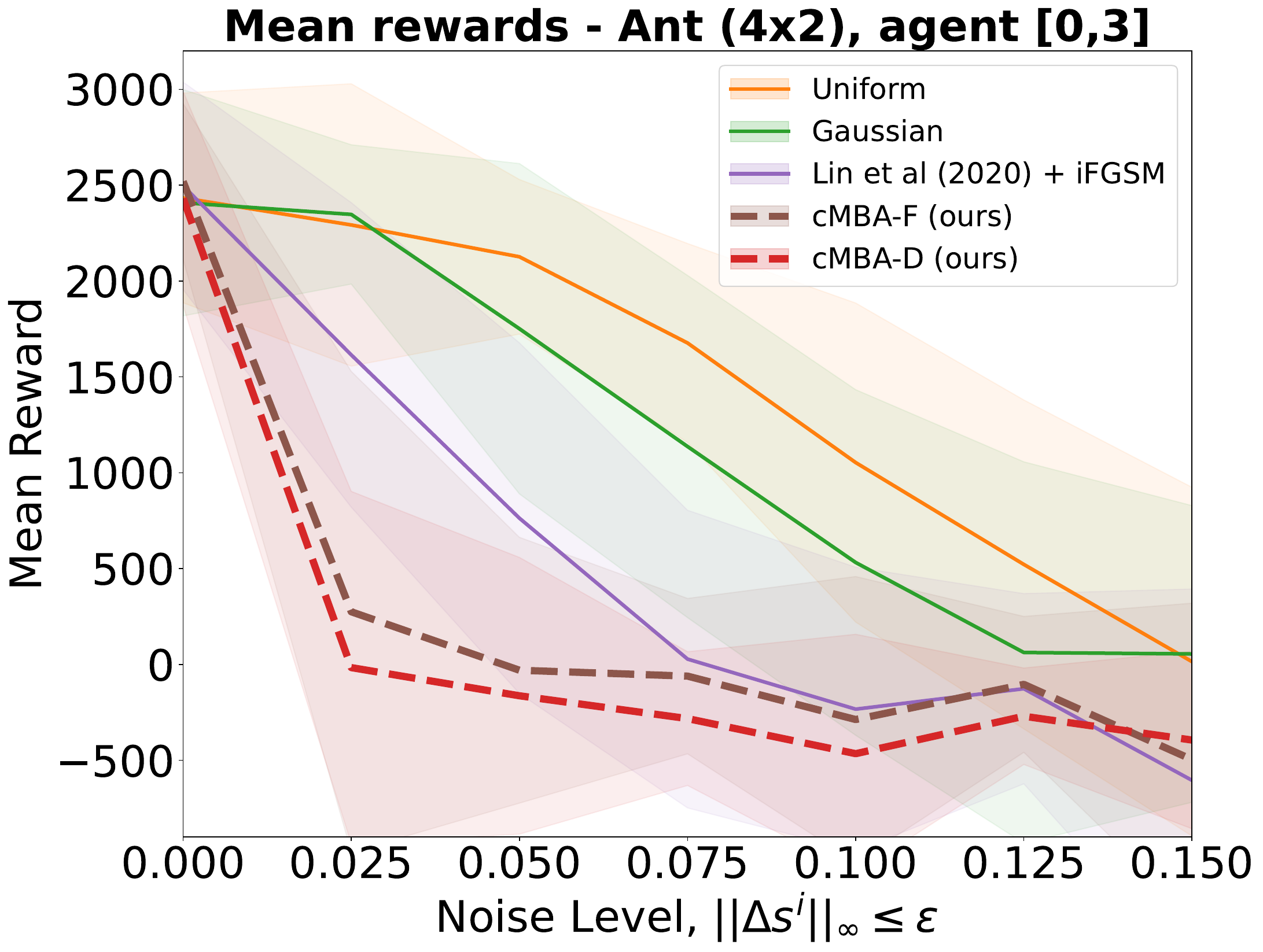}
    \includegraphics[width=0.33\linewidth,valign=t]{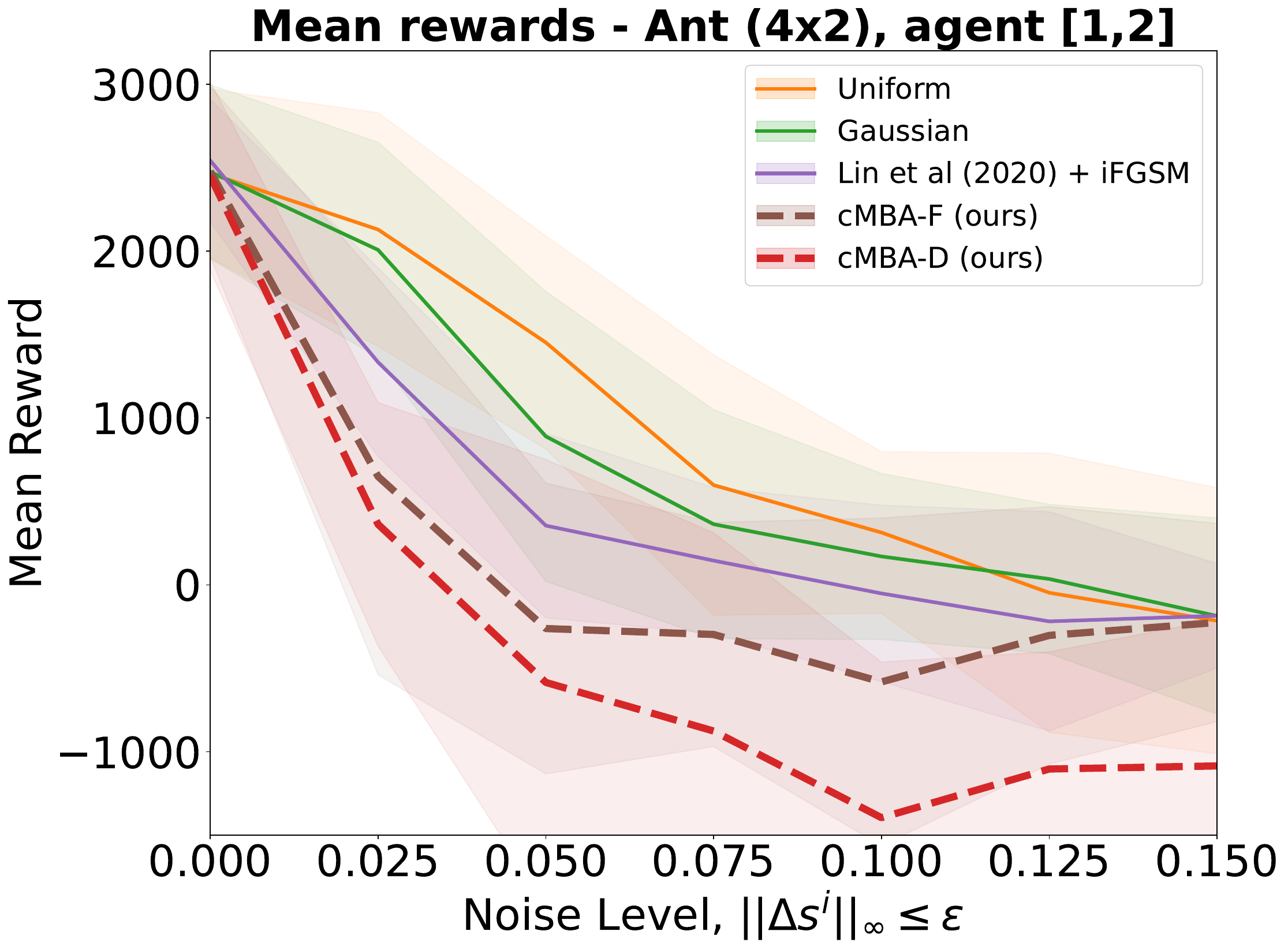}
    \includegraphics[width=0.33\linewidth,valign=t]{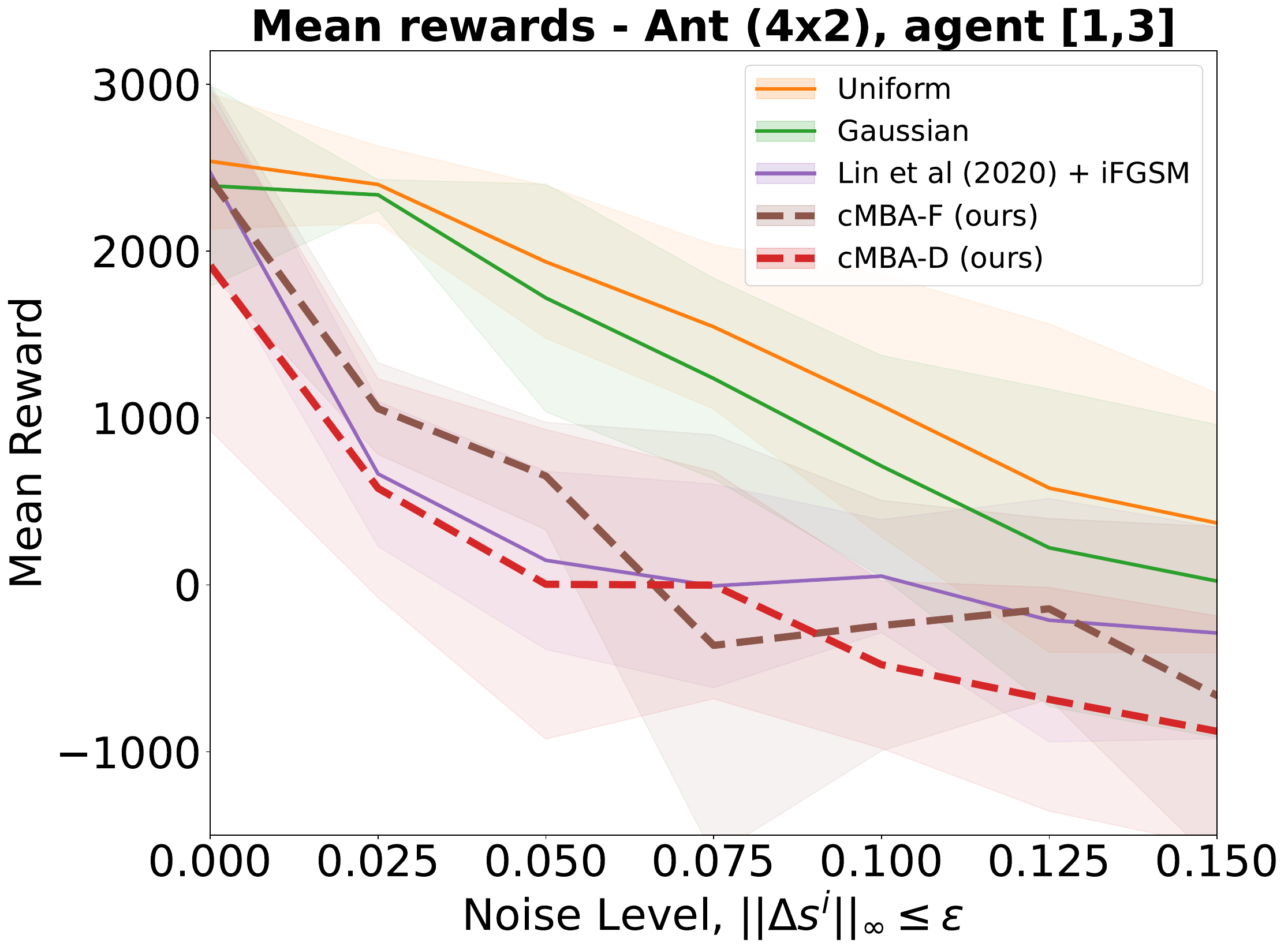}
    \includegraphics[width=0.33\linewidth,valign=t]{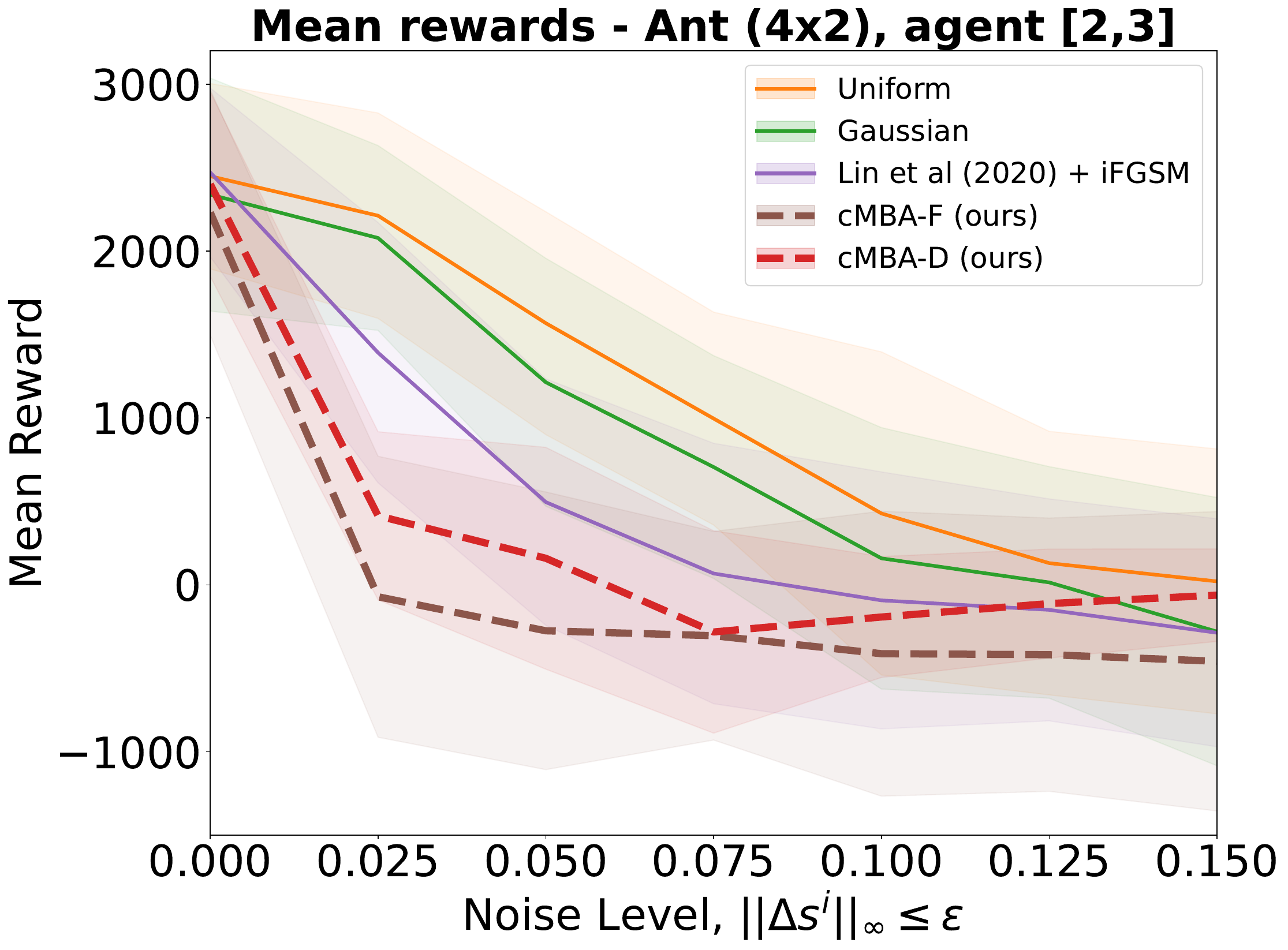}
    \caption{c-MBA vs baselines in \text{Ant(4x2)} when attacking two agents - \textbf{Exp. (III)}.}
    \label{fig:base_vs_model_ant_2agent_linf}
    \end{center}
    \vspace{-2ex}
\end{figure}

\textbf{Experiment (III) -- attacking two agents using model-free baselines vs model-based attack c-MBA using $\ell_{\infty}$ constrained:} We use model-free and model-based approaches to simultaneously attack two agents in \text{Ant(4x2)} environment. Fig.~\ref{fig:base_vs_model_ant_2agent_linf} illustrate the performance of various attacks. We observe that our c-MBA attack outperforms other baselines. For instance, at $\varepsilon=0.025$, the team reward reduction of c-MBA is 68\%, 522\%, 713\% \textit{more} than the three model-free baselines. In addition, our c-MBA-D even outperforms c-MBA-F to achieve lower reward at all budget levels.

\begin{figure}[ht!]
    \begin{center}
    \includegraphics[width=0.24\linewidth,valign=t]{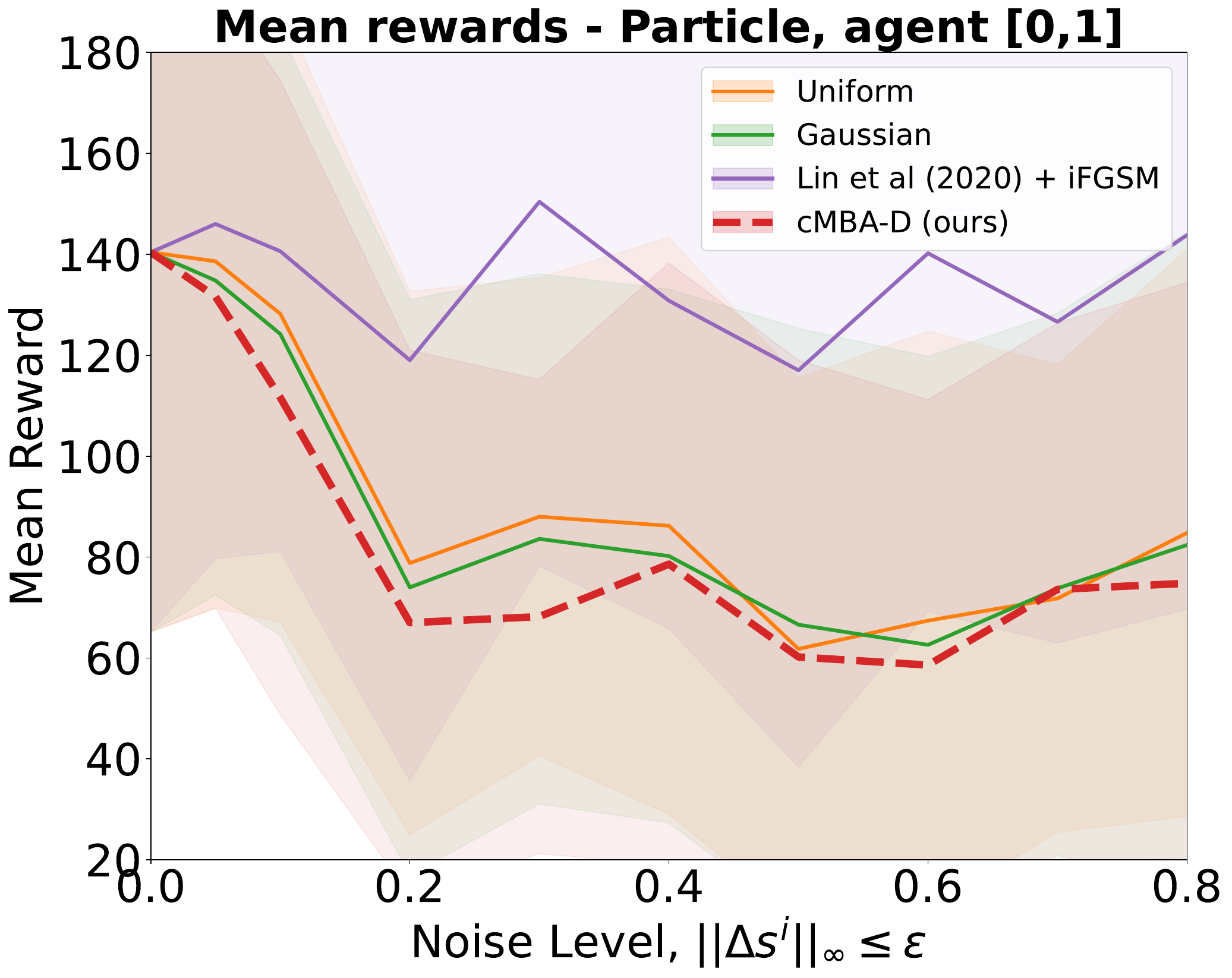}
    \includegraphics[width=0.24\linewidth,valign=t]{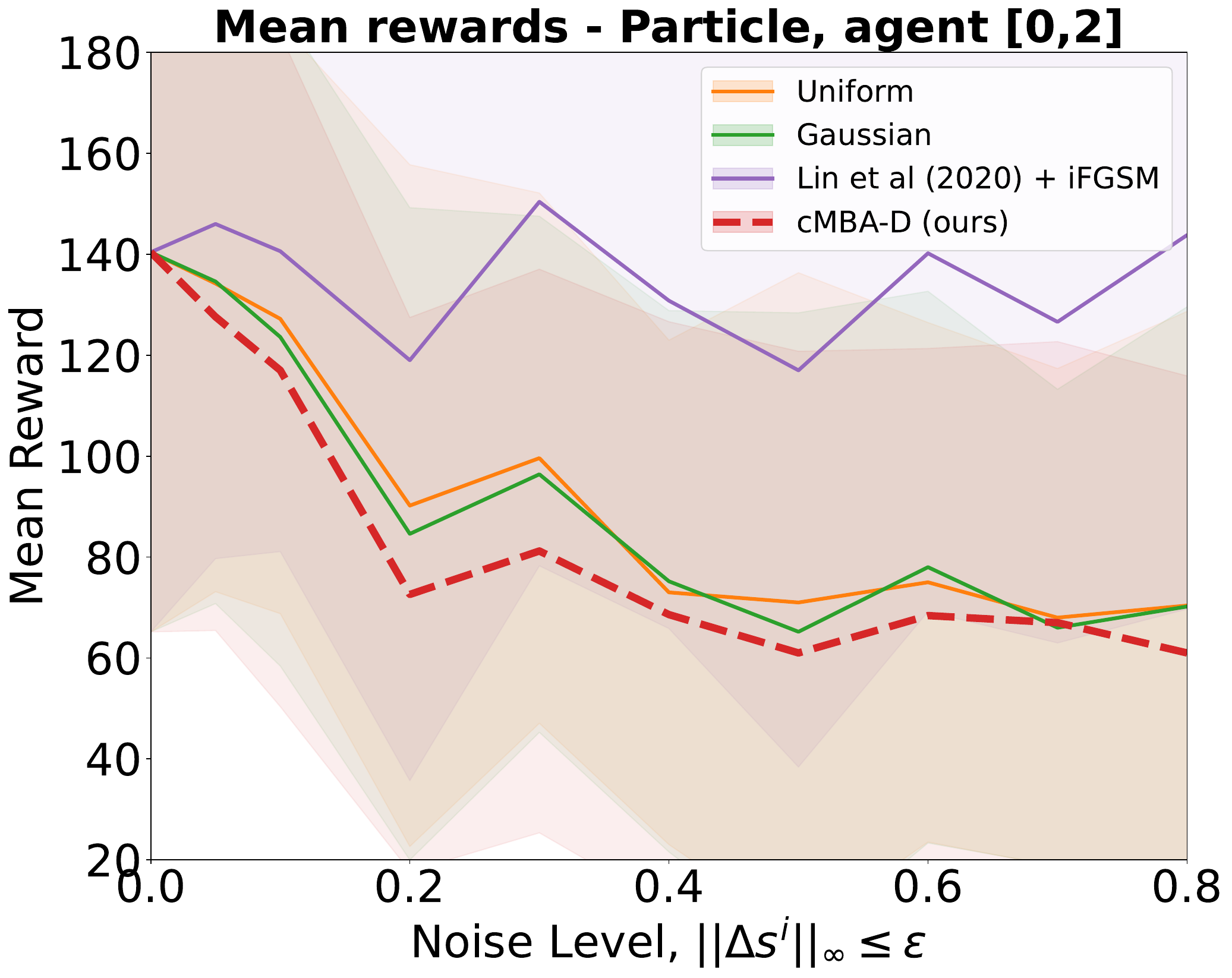}
    \includegraphics[width=0.24\linewidth,valign=t]{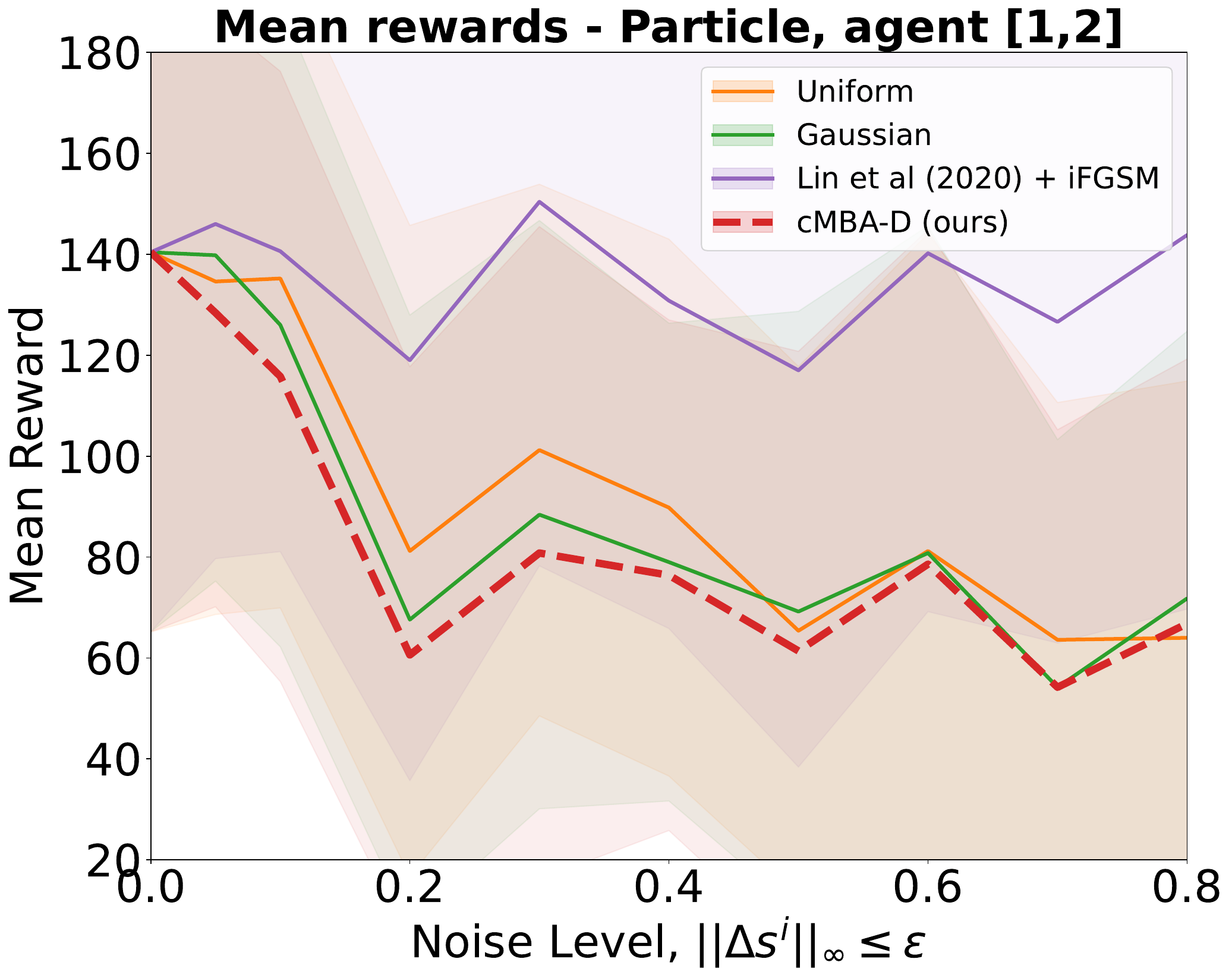}
    \includegraphics[width=0.24\linewidth,valign=t]{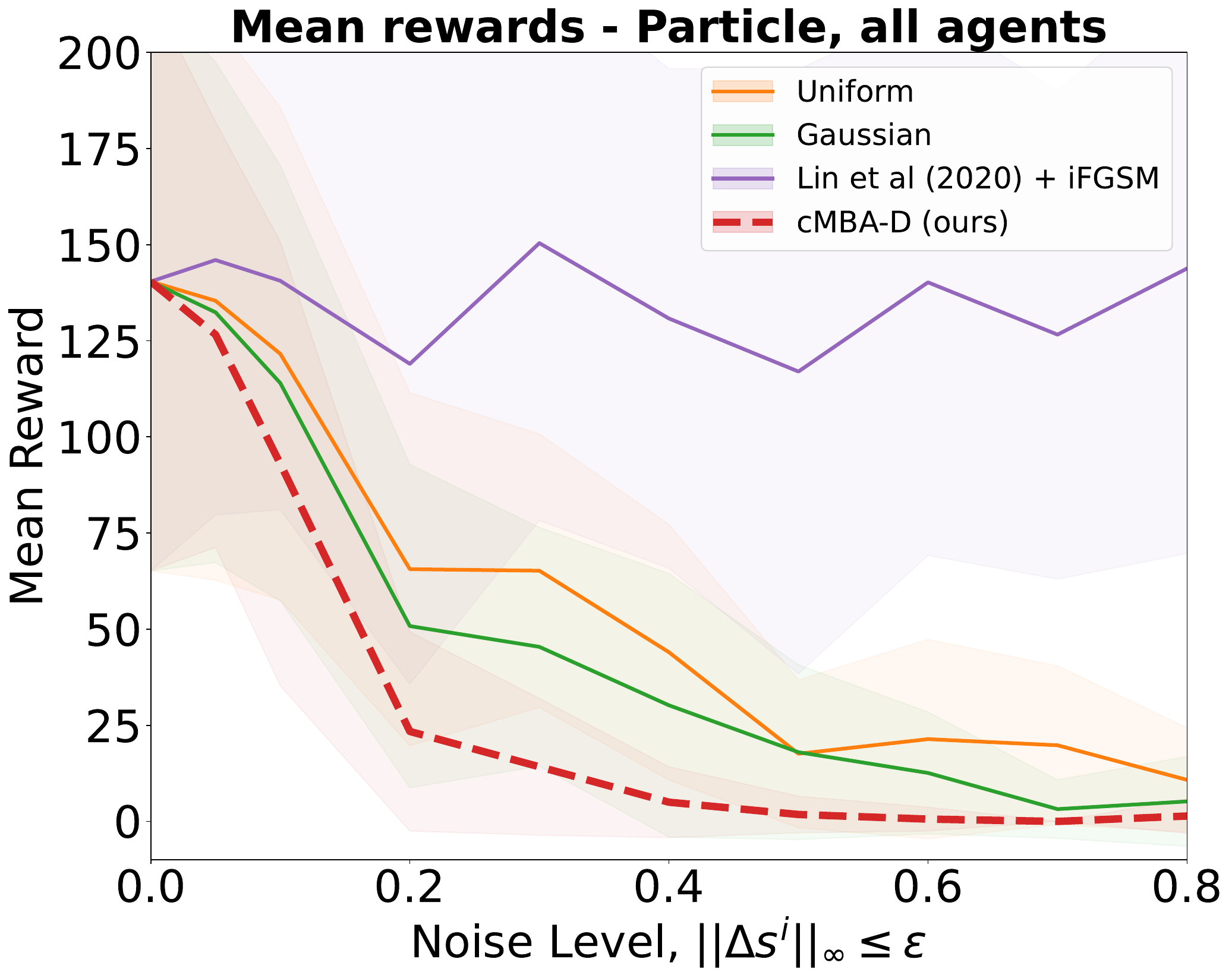}
    \caption{c-MBA vs baselines in \textbf{MPE(3x5)} when attacking two agents (3 leftmost) and three agents (rightmost) simultaneously using $\ell_\infty$ constraint - \textbf{Exp. (IV)}.}
    \label{fig:base_vs_model_particle_linf}
    \end{center}
    \vspace{-2ex}
\end{figure}

\textbf{Experiment (IV) -- model-free baselines vs model-based attack c-MBA in MPE(3x5) environment}: We perform the same procedure as in Experiment (III) to attack different agents in the \textbf{MPE(3x5)} environments where we attack two or three agents at the same time. Since we do not have expert knowledge about the failure state in this environment, we compare \textbf{c-MBA-D} with other model-free baselines. The results are depicted in Figure~\ref{fig:base_vs_model_particle_linf} where \textbf{c-MBA-D} perform slightly better than other random noise baselines when attacking 2 agents and significantly outperforms other baselines when attacking three agents simultaneously. We note that \citep{lin2020robustness} does not perform well in this experiment as we observe during training the adversarial policy that it could not lower the team reward effectively.

\begin{figure}[ht!]
    \begin{center}
    \includegraphics[width=0.24\linewidth,valign=t]{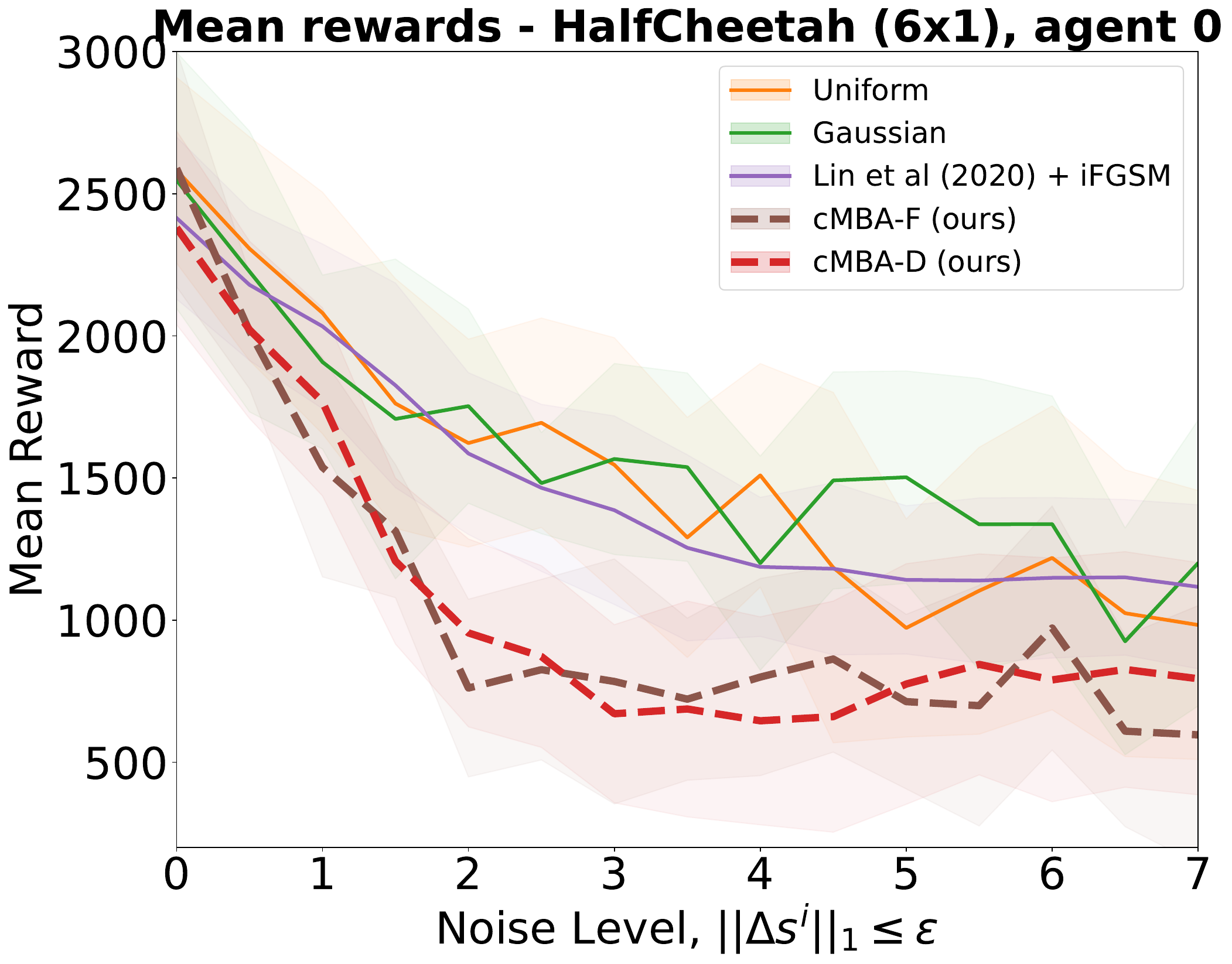}
    \includegraphics[width=0.24\linewidth,valign=t]{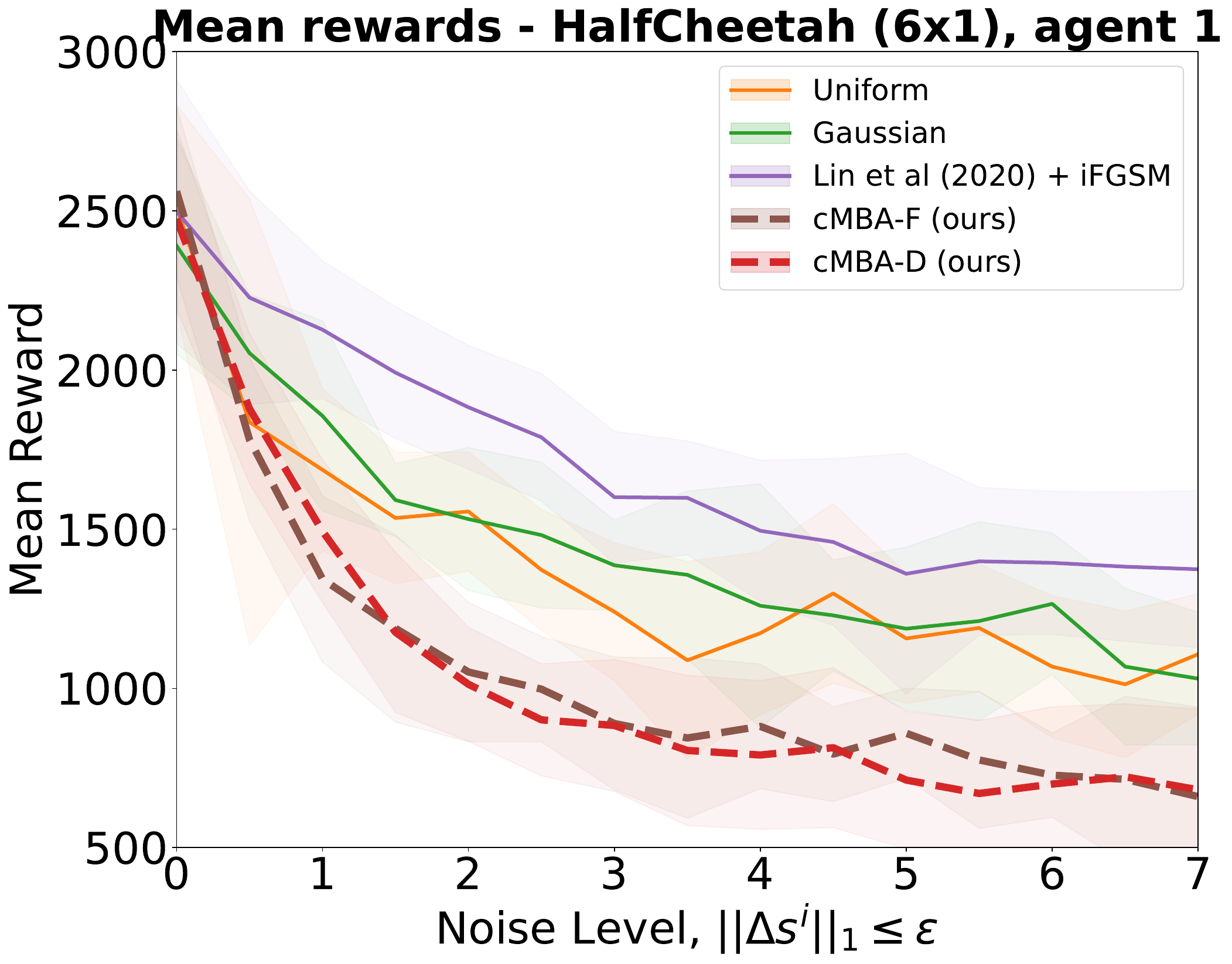}
    \includegraphics[width=0.24\linewidth,valign=t]{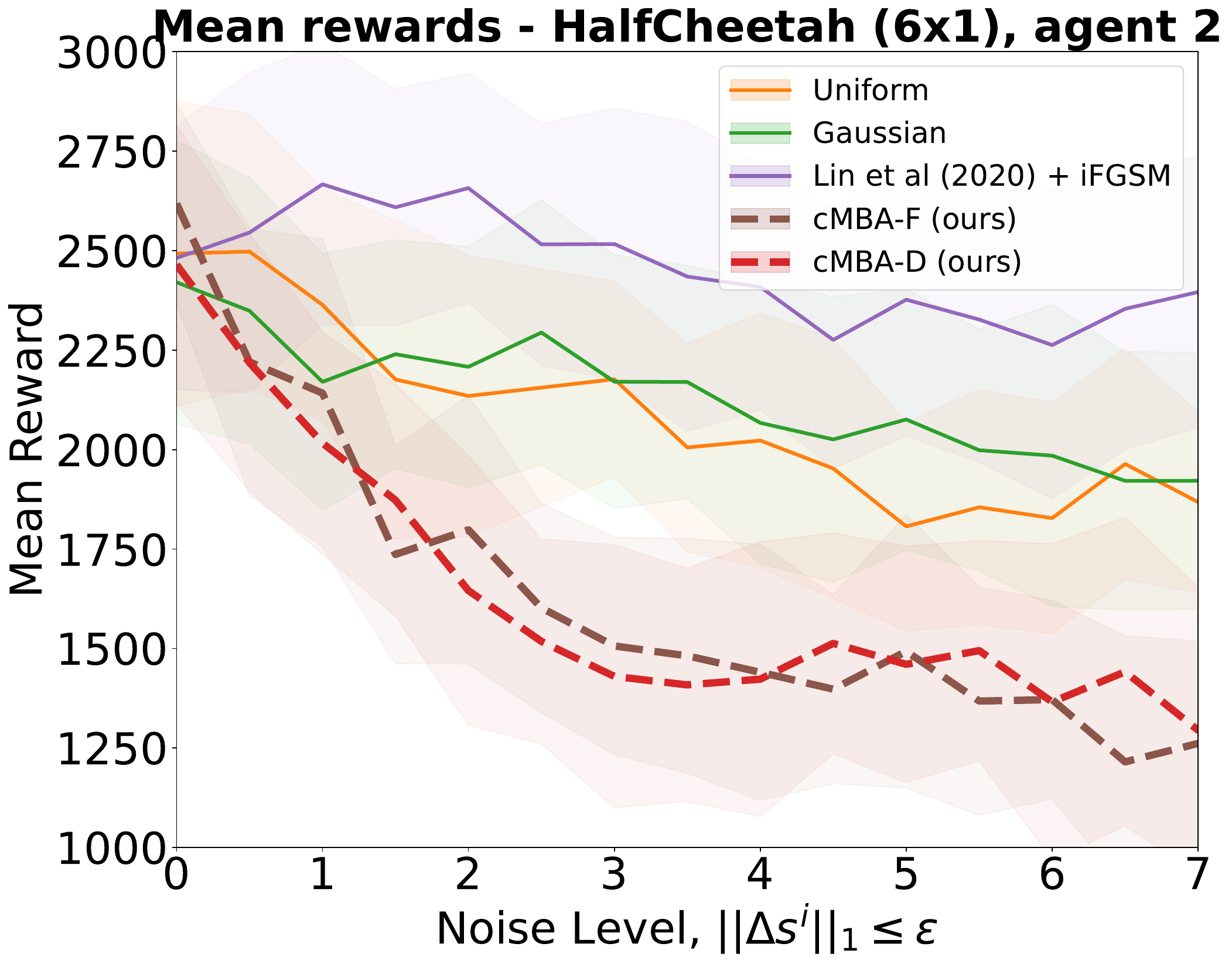}
    \includegraphics[width=0.24\linewidth,valign=t]{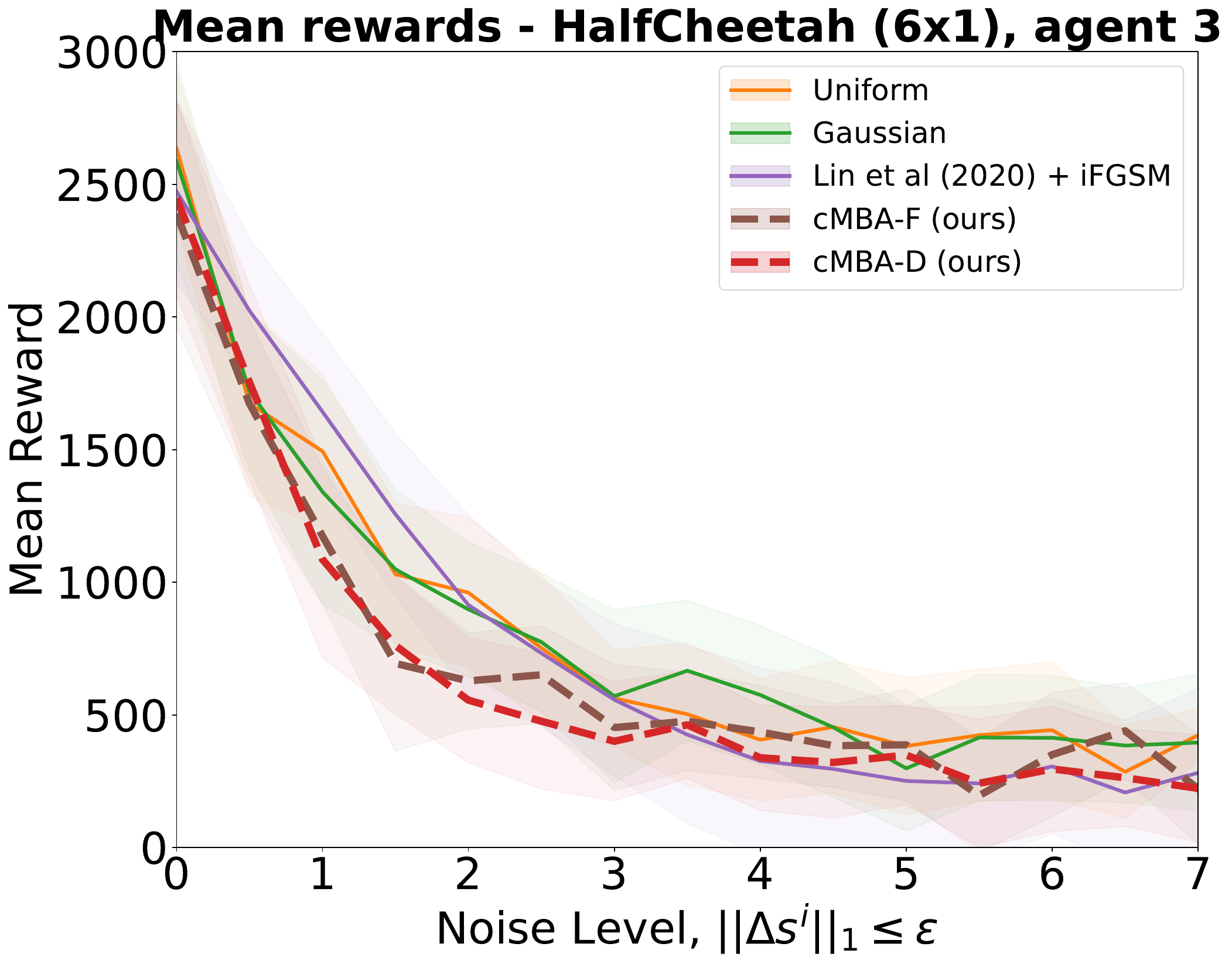}
    \includegraphics[width=0.24\linewidth,valign=t]{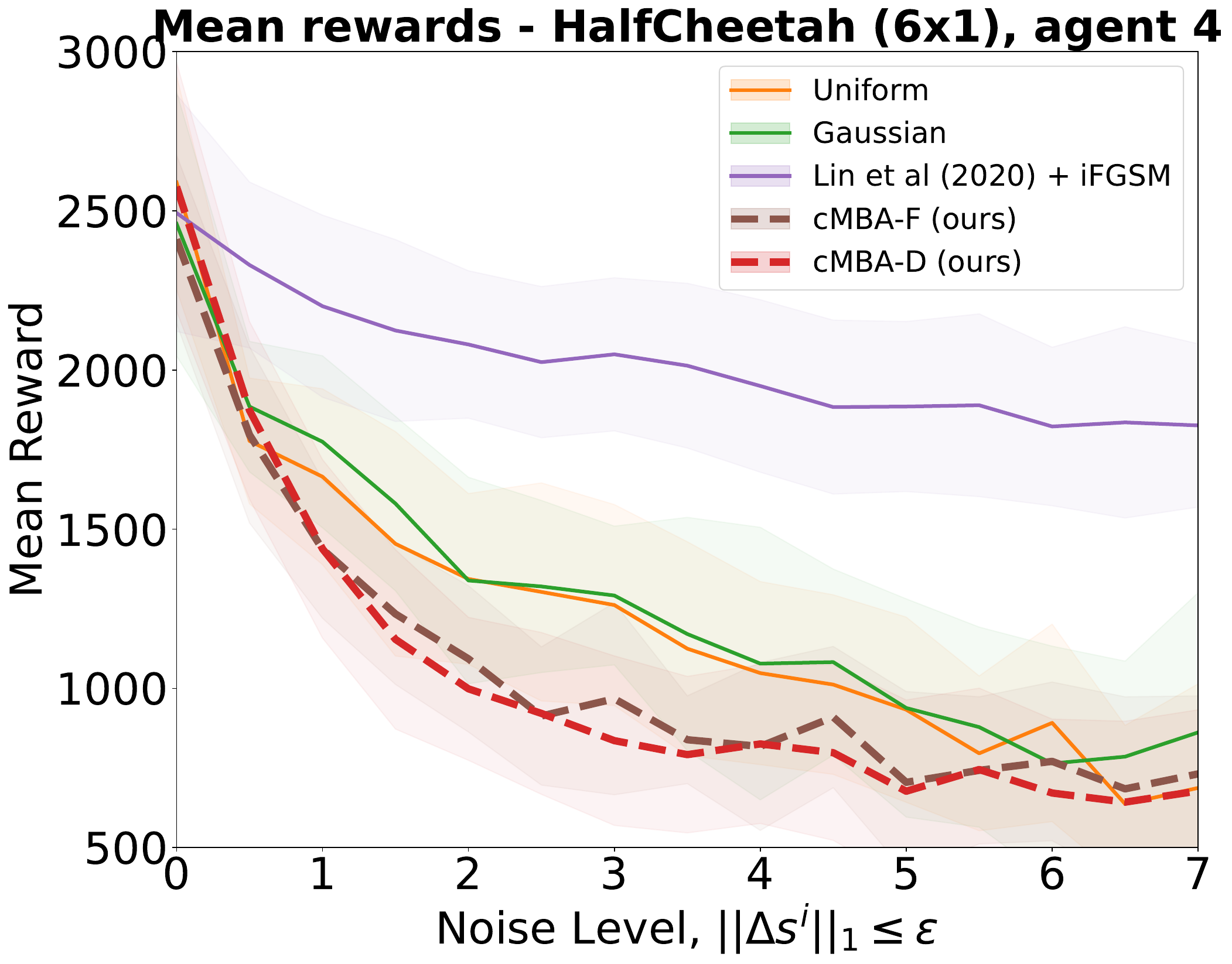}
    \includegraphics[width=0.24\linewidth,valign=t]{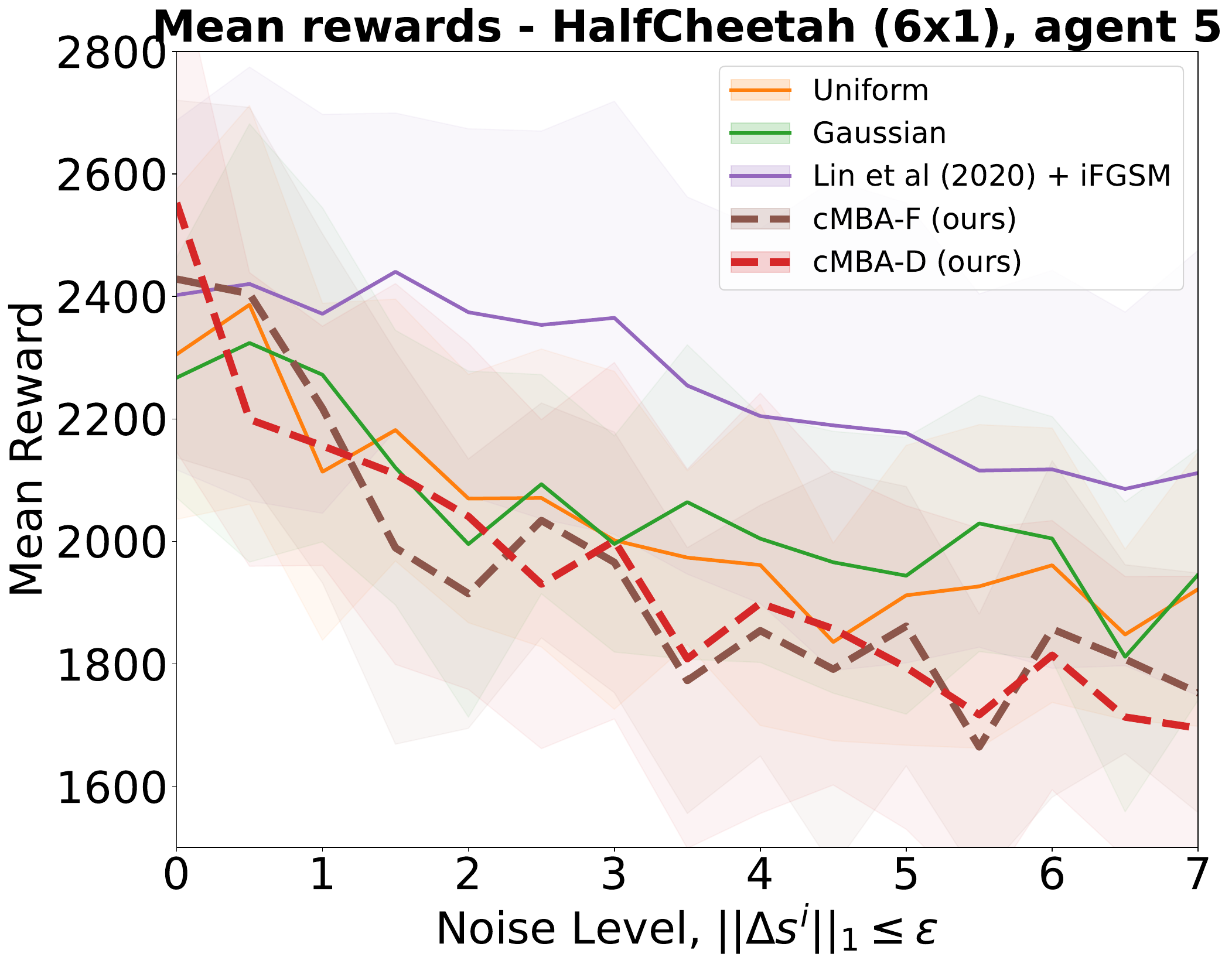}
    \includegraphics[width=0.24\linewidth,valign=t]{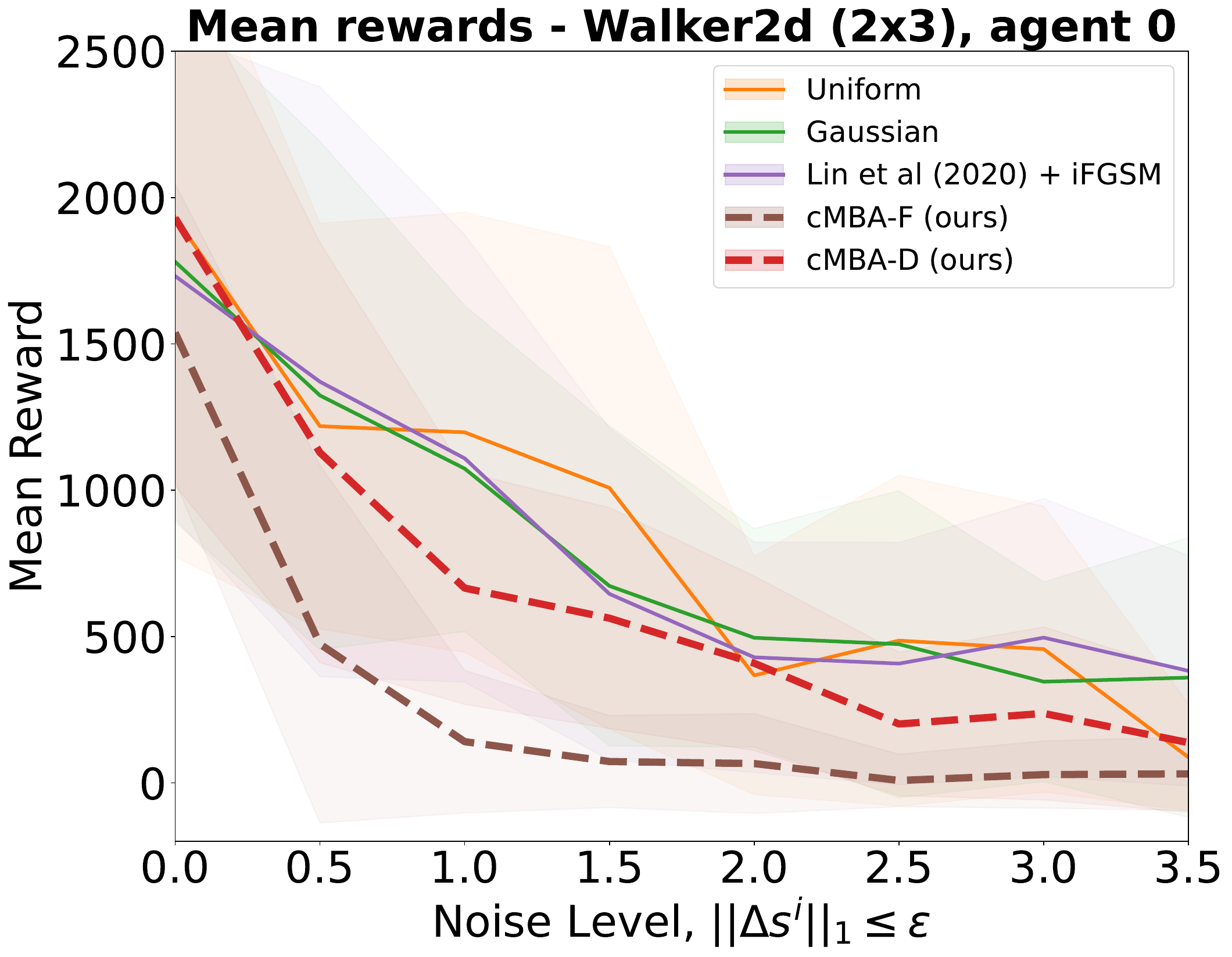}
    \includegraphics[width=0.24\linewidth,valign=t]{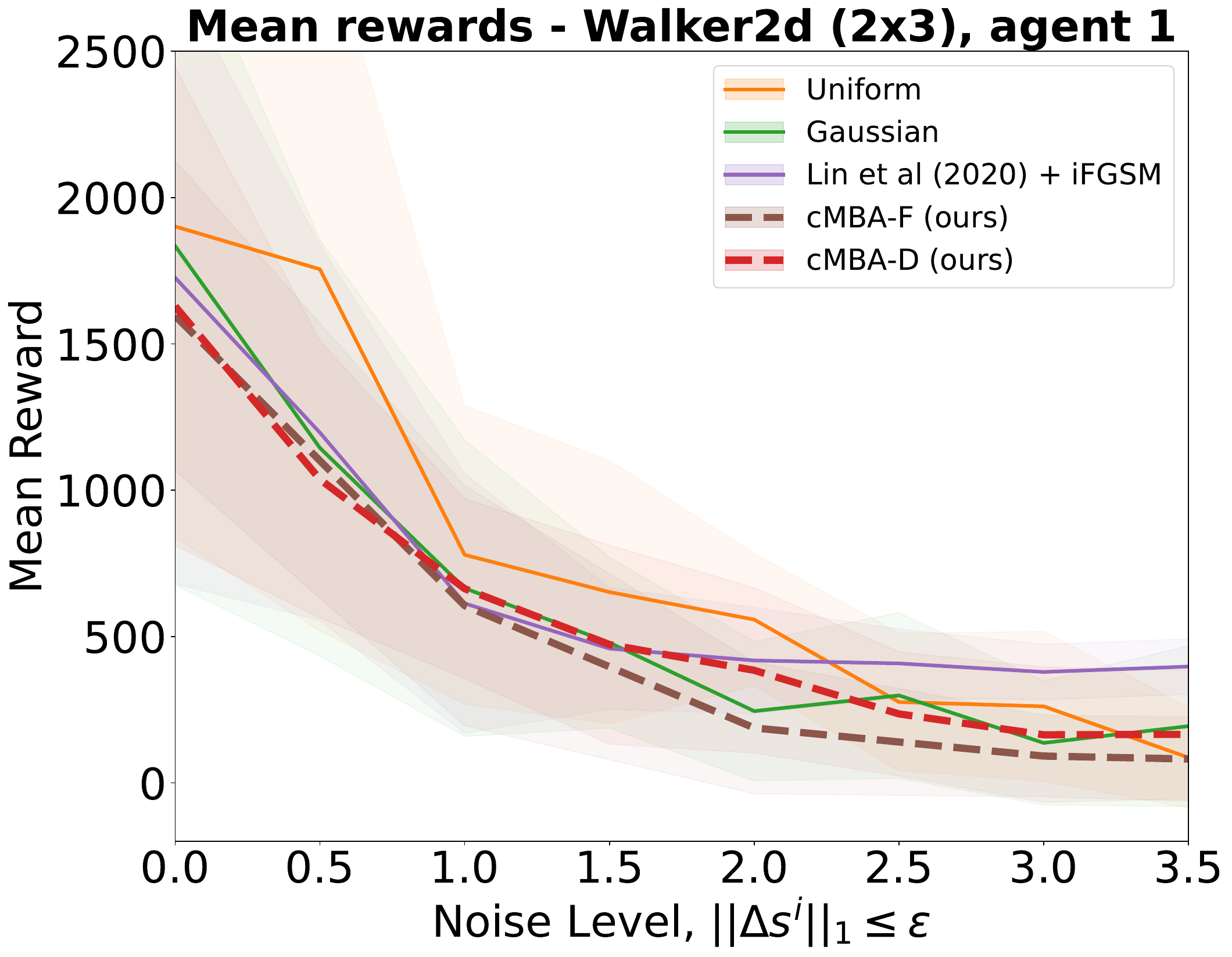}
    \caption{c-MBA vs baselines in \textbf{HalfCheetah(6x1)} and \textbf{Walker2d(2x3)} - \textbf{Exp. (V)}.}
    \label{fig:base_vs_model_cheetah_6x1_walker_l1_full}
    \end{center}
    \vspace{-3ex}
\end{figure}

\textbf{Experiment (V): model-free baselines vs model-based attack c-MBA on $\ell_{1}$ perturbation.} In addition to the $\ell_\infty$-norm budget constraint, we also evaluate adversarial attacks using the $\ell_1$-norm constraint. Note that using $\ell_1$-norm for budget constraint is more challenging as the attack needs to distribute the perturbation across all observations while in the $\ell_\infty$-norm the computation of perturbation for individual observation is independent. Fig.~\ref{fig:base_vs_model_cheetah_6x1_walker_l1_full} illustrate the effect of different attacks on \textbf{HalfCheetah(6x1)} and \textbf{Walker2d(2x3)} environments, respectively. Our c-MBA is able to outperform other approaches in almost all settings. In \textbf{HalfCheetah(6x1)}, using $\varepsilon=1.0$ under $\ell_1$ budget constraint, the amount of total team reward reduced by c-MBA variants is up to 156\%, 37\%, and 42\% \textit{more} than Lin et al. (2020) + iFGSM, Gaussian, and Uniform baselines, respectively.


\begin{figure}[t!]
    \begin{center}
    \includegraphics[width=0.42\linewidth,valign=t]{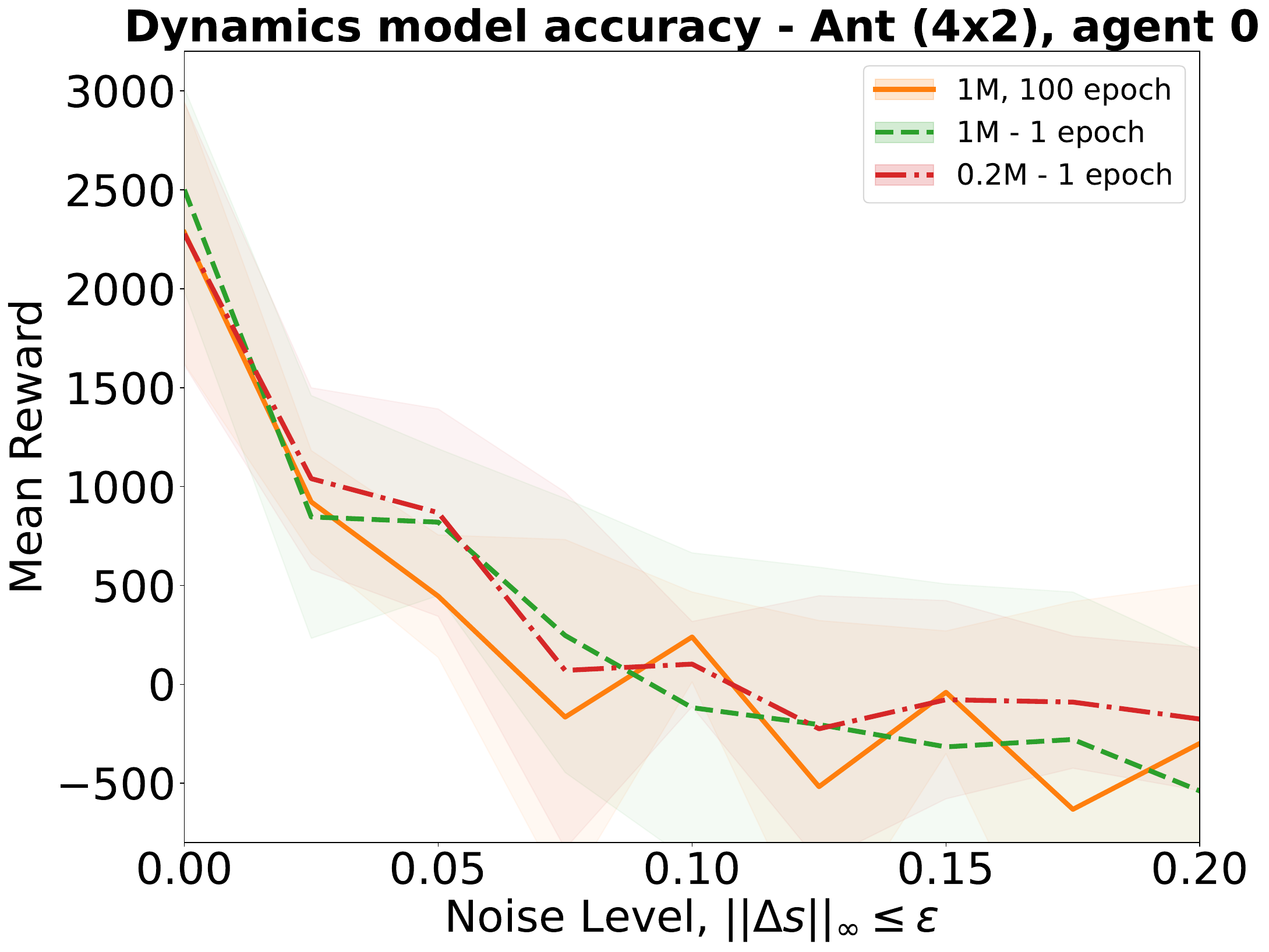}
    \includegraphics[width=0.42\linewidth,valign=t]{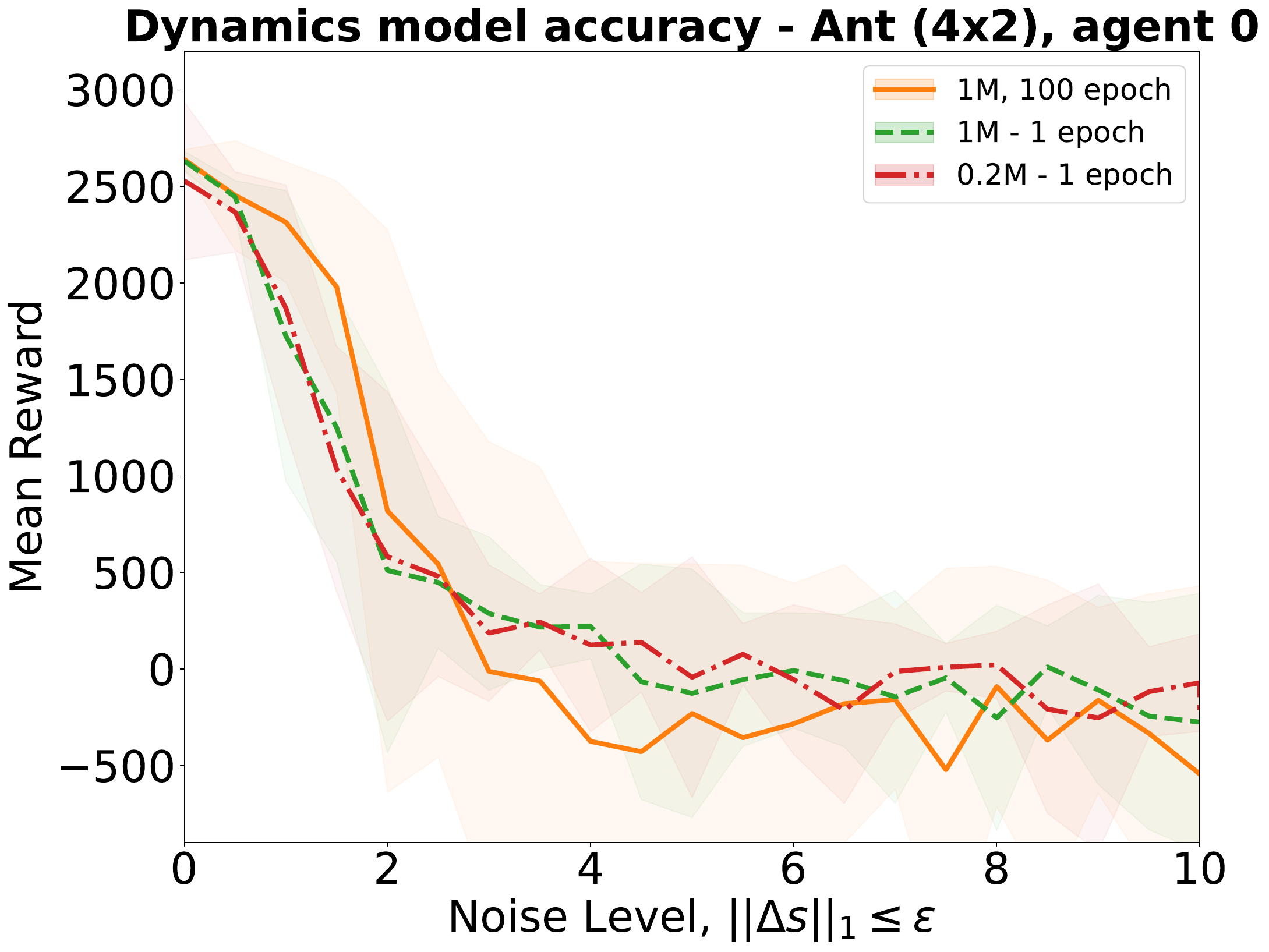}
    \caption{{Effect of using three dynamics models in \textbf{Ant(4x2)} when attacking Agent $0$ out of four agents in \textbf{Ant(4x2)} - \textbf{Exp. (VI)}.}
    }
    \label{fig:ant_model_acc}
    \end{center}
\end{figure}

\textbf{Experiment (VI) -- adversarial attacks using dynamics model with various accuracy.} We compare our attack when using trained dynamics model with different mean-squared error (MSE). We use \textbf{Ant(4x2)} environment and consider the following three dynamics models: 
\begin{itemize}\setlength\itemsep{-0.075ex}
    \item \textbf{model(1M, 100 epochs)} is trained using 1 million samples for 100 epochs and select the model with the best performance.
    \item \textbf{model(1M, 1 epoch)} is trained using 1 million samples for only 1 epoch.
    \item \textbf{model(0.2M, 1 epoch)} is trained using less number of samples to only 200k and we train it for only 1 epoch. 
\end{itemize}
These models are evaluated on a predefined test set consisting of 0.1M samples. The test MSE of these models are 0.33, 0.69, and 0.79, respectively, with the initial (w/o training) test MSE of 1.241. Fig.~\ref{fig:ant_model_acc} depicts the attacks using these three models on the same agent in Ant(4x2) environment using $\norm{\cdot}_{\infty}$ budget constraint. Interestingly, the dynamics model trained with only 0.2M samples for 1 epoch can achieve comparable performance with the other two models using 5$\times$ more samples.

\begin{figure*}[htp!]
    \begin{center}
    \includegraphics[width=0.8\linewidth,valign=t]{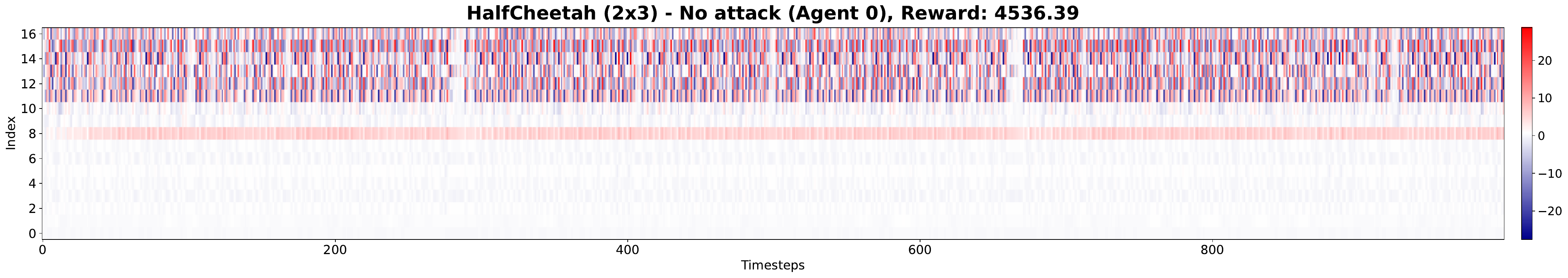}
    \includegraphics[width=0.8\linewidth,valign=t]{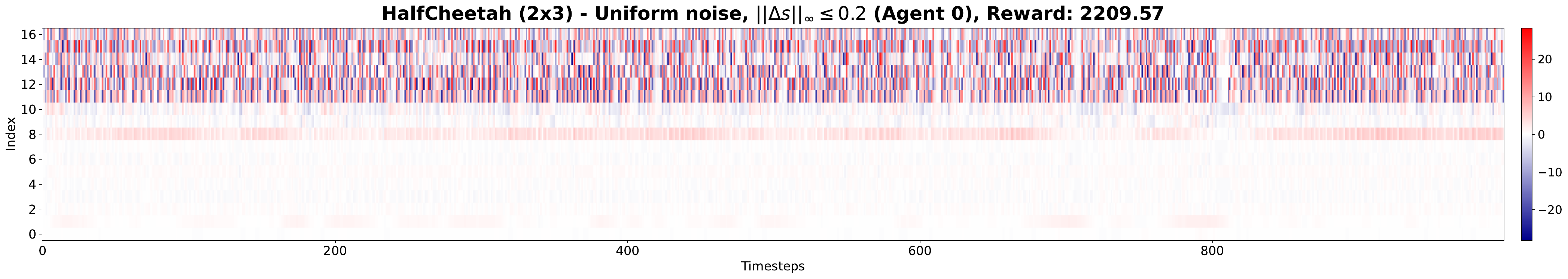}
    \includegraphics[width=0.8\linewidth,valign=t]{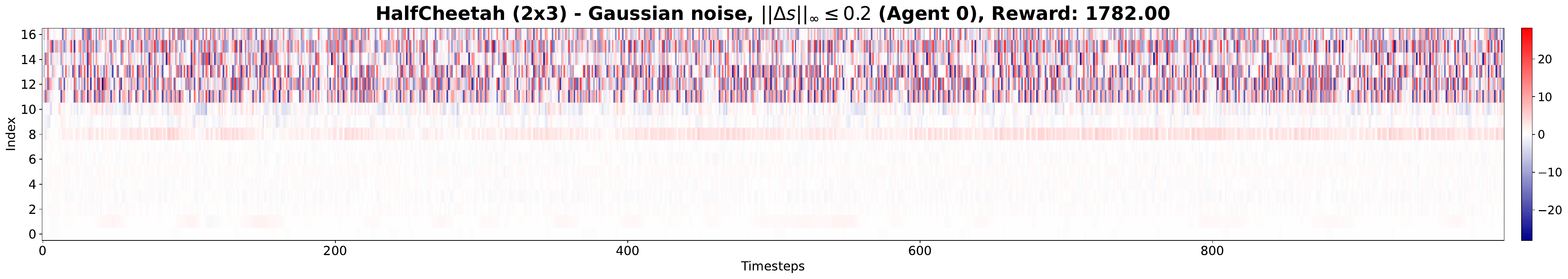}
    \includegraphics[width=0.8\linewidth,valign=t]{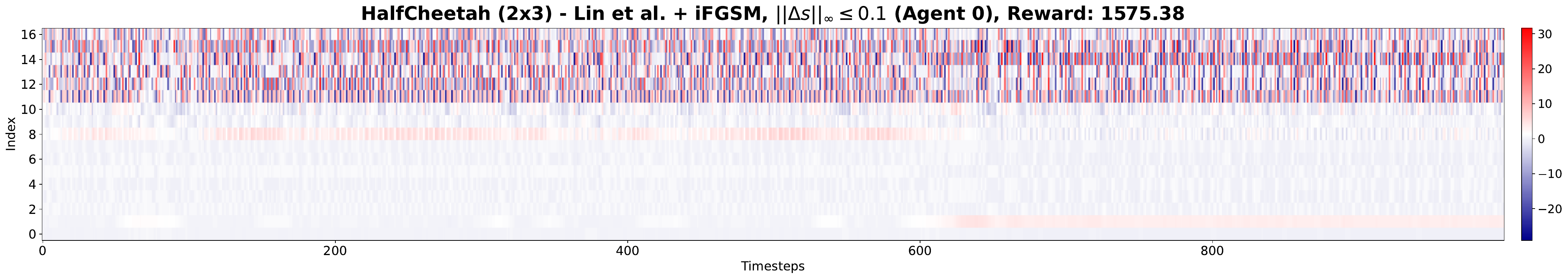}
    \includegraphics[width=0.8\linewidth,valign=t]{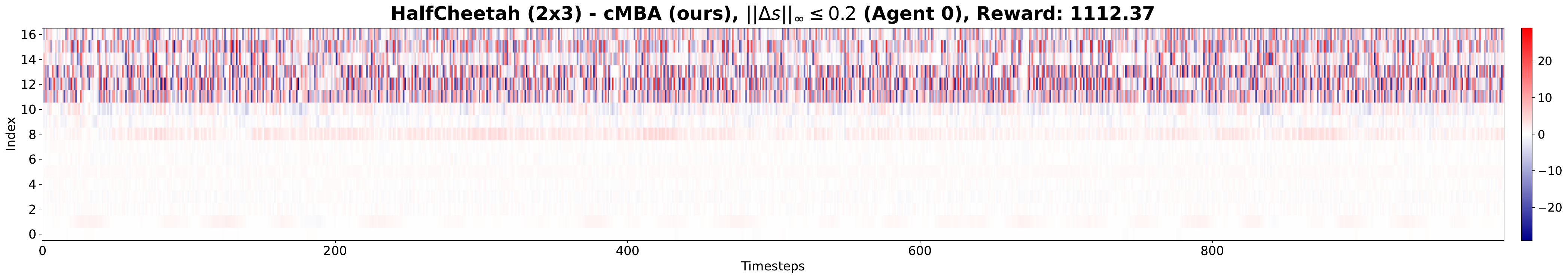}
    \caption{Recordings of state values in an episode under different attacks one out of two agents (Agent $0$) in \textbf{HalfCheetah( 2x3)} environment.}
    \label{fig:cheetah_2x3_linf_state_val}
    \end{center}
\end{figure*}

In addition to the visualization in Fig.~\ref{fig:ant_linf_render}, we investigate how the state values change during these attacks. Fig.~\ref{fig:cheetah_2x3_linf_state_val} presents different recordings of state values under adversarial attacks compared to no attack. Consider \textbf{state index $\boldsymbol{8}$}, which represents the \textbf{horizontal velocity} of the agent. For the \textbf{HalfCheetah(2x3)} environment, as the goal is to make the agent move forward as fast as possible, we expect the reward to be proportional to this state value. From Fig.~\ref{fig:cheetah_2x3_linf_state_val}, all three attacks have fairly sparse red fractions across time-steps, which result in a much lower reward compared to the no-attack setting. Among the three attacks, our model-based attacks have the most sparse red fractions leading to the lowest rewards. In addition, the model-based attack appears to show its advantage in environments with more agents as our approach results in lower rewards under a smaller budget as seen in \textbf{HalfCheetah(6x1)} and \textbf{Ant(4x2)} environments. 


In summary, our c-MBA attack is able to shows its advantage in all tested multi-agent environment where it achieves lower reward with smaller budget level. Moreover, c-MBA with the victim agent selection has been shown to constantly performs better than the original c-MBA variant as seen in Figure~\ref{fig:model_variants}.

\section{Conclusions}\label{sec:conclusion}

We propose a new attack algorithm named \textbf{c-MBA} for evaluating the robustness of c-MARL environment. Our algorithm is the first model-based attack to craft adversarial observation perturbations and we have shown that c-MBA outperforms existing model-free attack baselines by a large margin 
under both MA-MuJoCo and MPE benchmarks. Unique to multi-agent setting, we also propose a new victim-selection strategy to select the most vulnerable victim agents given the attack budget, which has not been studied in prior work. We show that with the victim-agent selection strategy, c-MBA attack becomes stronger and more effective. We also propose the first data-driven approach to determine the failure state based on the pre-collected data without extra overhead making our attack more flexible.
\bibliographystyle{plainnat}
\bibliography{ref}





\end{document}